%% file: main.tex
\documentclass[runningheads]{llncs}

\usepackage{eccv}

\usepackage{eccvabbrv}

\usepackage{graphicx}
\usepackage{booktabs}
\usepackage{multirow}
\usepackage{pifont}
\usepackage{adjustbox}
\usepackage{tcolorbox}
\tcbuselibrary{listings}
\usepackage{varwidth}
\usepackage{fancyvrb}
\usepackage{wrapfig}
\usepackage{subcaption}
\usepackage{caption}
\usepackage{paralist}
\usepackage{algorithm}
\usepackage{algpseudocode}
\usepackage[accsupp]{axessibility}  % Improves PDF readability for those with disabilities.

\usepackage{xcolor}

\usepackage[pagebackref,breaklinks,colorlinks,citecolor=eccvblue]{hyperref}
\usepackage{orcidlink}

\begin{document}

\title{PrintAnything: Learning Geometric Plan Map \\for 3D Printing G-code Generation \\from Unoriented Point Clouds} 

\titlerunning{PrintAnything}

% Second Author\inst{2,3}\orcidlink{1111-2222-3333-4444} \and
% Third Author\inst{3}\orcidlink{2222--3333-4444-5555}}

% % TODO FINAL: Replace with an abbreviated list of authors.

% % TODO FINAL: Replace with your institution list.
% Springer Heidelberg, Tiergartenstr.~17, 69121 Heidelberg, Germany

\author{Sangmin Hong\inst{1}\orcidlink{0009-0009-1881-3548} \and
Daniel Sungho Jung\inst{1}\orcidlink{0000-0002-9955-418X} \and
Heewon Kim\textsuperscript{$\dagger$}\inst{3,4}\orcidlink{0000-0001-7777-9823} \and
Kyoung Mu Lee\textsuperscript{$\dagger$}\inst{1,2}\orcidlink{0000-0001-7210-1036}}

\authorrunning{S.~Hong et al.}

\institute{
IPAI, Seoul National University, Seoul, Korea \and
Dept. of ECE \& ASRI, Seoul National University, Seoul, Korea \and
Soongsil University, Seoul, Korea \and
Kairoba Inc.\\[0.5ex]
\email{\{mchiash2,dqj5182,kyoungmu\}@snu.ac.kr, hwkim@ssu.ac.kr}
}

\maketitle

\renewcommand{\thefootnote}{}%
\footnote{$\dagger$ co-corresponding author.}%
\renewcommand{\thefootnote}{\arabic{footnote}}%

\input{sec/0_abstract}

\input{sec/1_introduction}
\input{sec/2_related_works}
\input{sec/3_proposed_method}

\input{sec/4_implementation_details}
\input{sec/5_experiments}
\input{sec/6_conclusion}
\input{sec/acknowledgement}

% \clearpage


\clearpage  % TODO FINAL: This \clearpage needs to be removed from both review and camera-ready versions.

% \section*{Acknowledgements}
% Please insert your acknowledgments here.

\bibliographystyle{splncs04}
\bibliography{main}
\clearpage
\input{sec/X_suppl}
\end{document}

%% file: sec/0_abstract.tex
\begin{abstract}
Point clouds are one of the most fundamental and widely used 3D representations, serving as the most basic geometric representation of 3D shapes.
Nevertheless, most existing 3D printing pipelines require a watertight mesh as input, preventing the direct use of point clouds for fabrication.
A common workaround is to reconstruct meshes from point clouds; however, the resulting meshes often contain geometric artifacts, such as incorrect faces or topological inconsistencies, that are difficult to repair and may lead to printing failures.
To overcome these limitations, we propose PrintAnything, a novel framework that learns to produce executable 3D printing G-code directly from 3D point clouds without requiring mesh reconstruction.
To enable point clouds to serve as direct input for slice-wise toolpath generation, we introduce a slice-wise point projection strategy that transforms unstructured 3D point clouds into slice-aligned 2D representations consistent with layer-by-layer nature of fused deposition modeling in 3D printing.
To eliminate mesh dependency and provide a unified representation that bridges point clouds and G-code, we propose Geometric plan~(G-plan) map, a compact 2D representation composed of occupancy, region, and flow maps that encode the geometric and extrusion properties required for toolpath synthesis in 3D printing.
As a result, our proposed method accurately generates printable G-code directly from point clouds, enabling a practical and fully mesh-free pipeline for 3D printing.
The code is publicly available at \href{https://github.com/Sangminhong/PrintAnything}{https://github.com/Sangminhong/PrintAnything}.

  \keywords{3D Printing \and Manufacturing \and G-code \and Point cloud}
\end{abstract}

%% file: sec/1_introduction.tex
\section{Introduction}
\label{sec:intro}

\input{fig/figure1}

Point clouds serve as a fundamental representation of 3D geometry and frequently appear as the native output of modern sensors and cameras.
They are naturally produced by LiDAR sensors~\cite{hong2023acl}, RGB-D cameras~\cite{hong2025starrygazer}, and increasingly by 3D generative models~\cite{kim2021setvae, sanghi2022clip, vahdat2022lion, huang2024pointinfinity, ren2024tiger, zhou2024frepolad, mo2024fast, du2025superpc, lee2025rgb2point}.
At the same time, 3D printing has emerged as a widely adopted technology for fabricating 3D geometry due to its accessibility and effectiveness for rapid prototyping and customized manufacturing.
However, the integration of point cloud and 3D printing remains limited in practice.
Consequently, enabling the direct fabrication of point clouds through 3D printing presents an important yet underexplored problem.

There are two major issues that prevent point clouds from being directly used for 3D printing.
First, point clouds cannot be directly taken as input by existing 3D printing pipelines.
In typical fabrication workflows, a digital 3D model is converted into slice-wise toolpaths using slicing software~\cite{prusaslicer}, and these toolpaths are executed through G-code instructions that specify nozzle motion and material extrusion.
These pipelines assume a watertight mesh as the input representation, making point clouds incompatible with standard slicing tools even though they are often the most readily available 3D signal in many acquisition pipelines.
Second, converting point clouds into meshes can introduce errors that lead to printing failures.
A common practice is to reconstruct meshes from point clouds~\cite{kazhdan2006poisson,hanocka2020point2mesh, peng2021shape,liu2025diffusing, chenmeshanything} before slicing, but reconstructing surfaces from sparse or noisy point clouds is inherently ill-posed and frequently produces artifacts such as incorrect faces, holes, or topological inconsistencies.
These defects are difficult to repair and can propagate into the slicing stage, resulting in invalid toolpaths or unstable extrusion during fabrication.
The problem is particularly severe for unoriented point clouds that consist only of raw point coordinates without normals as shown in Figure~\ref{fig:overview}.
Consequently, the dependence on mesh both prevents unoriented point clouds from serving as direct inputs and introduces potential failure propagation in the 3D printing pipeline for point clouds.

To overcome these challenges, we propose PrintAnything, a framework that learns to generate executable 3D printing G-code directly from point clouds without requiring mesh reconstruction.
To enable point clouds to be used as input for 3D printing pipelines, we introduce a slice-wise point projection strategy that converts an unstructured point cloud into slice-aligned 2D inputs encoding local geometric context around each slice height.
These slice-aligned representations allow the model to reason about the geometry of each printing layer while preserving information from the original 3D point cloud.
Furthermore, to eliminate failures caused by erroneous mesh reconstruction, we introduce the Geometric plan (G-plan) map, a compact slice-wise representation composed of an occupancy map, a region map, and a flow map.
The occupancy map specifies which spatial locations correspond to printed material, the region map encodes structural categories required for toolpath synthesis~(\textit{e.g.,} wall or infill), and the flow map represents extrusion-related properties for material deposition as illustrated in Figure~\ref{fig:overview}.
Based on this representation, PrintAnything predicts G-plan maps for each slice from the input point cloud, generates appropriate infill patterns within the predicted infill regions, and finally converts the resulting maps into ordered toolpaths to produce executable G-code.
Through this pipeline, PrintAnything enables a fully mesh-free approach that directly transforms point clouds into printable instructions for 3D fabrication.

In summary, our key contributions are as follows:
\begin{itemize}
    \item We introduce PrintAnything, a framework that generates executable 3D printing G-code directly from point clouds, enabling a practical and fully mesh-free fabrication pipeline.
    \item To align point-cloud geometry with slice-wise printing, we propose a slice-wise point projection strategy that converts unstructured 3D points into slice-aligned 2D representations suitable for toolpath-related learning.
    \item To bridge point clouds and G-code with a unified representation, we propose the Geometric plan (G-plan) map, a compact slice-wise representation composed of occupancy, region, and flow maps that encode both geometric and extrusion properties required for toolpath synthesis.
    \item As a result, our method accurately generates printable G-code directly from point clouds without mesh reconstruction, demonstrating a robust mesh-free pipeline for 3D printing.
\end{itemize}

%% file: fig/figure1.tex
\begin{figure}[t]
% \phantomsubcaption\label{fig:challenges:a}
% \phantomsubcaption\label{fig:challenges:b}
% \phantomsubcaption\label{fig:overview:c}
\begin{center}
    \includegraphics[width=\linewidth]{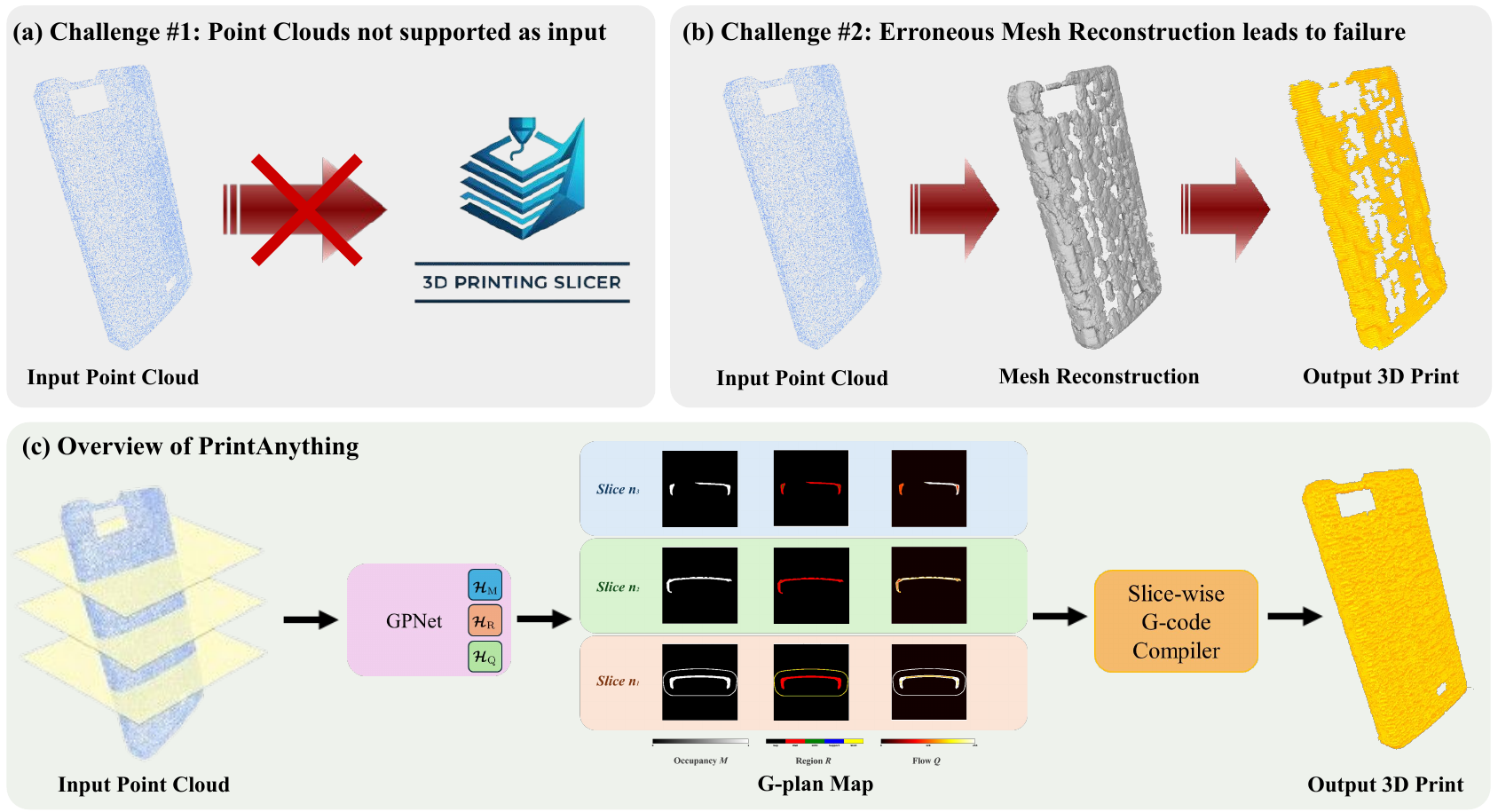}
    \end{center}
    \caption{
        \textbf{Motivated by Two Challenges:} (a) No Point Cloud support for 3D Printing Slicers (b) Point Cloud to Mesh reconstruction is limited and error-prone. We present \textbf{(c) PrintAnything} converts point clouds into a G-plan Map and compiles slice-wise G-code for printing.
    }
    \label{fig:overview}
\end{figure}

%% file: sec/2_related_works.tex
\section{Related works}
\label{sec:related_works}

\noindent\textbf{3D Printing.}
Recent advances in 3D printing have improved speed, accuracy, structural coherence, and robustness across diverse geometries~\cite{gueners2020stiffness, li2020slaam, munasinghe2021temperature, koyama2015autoconnect, zhou2016thingi10k, wu2016printing}.
Prior work has explored printable connectors~\cite{koyama2015autoconnect}, large-scale model datasets~\cite{zhou2016thingi10k}, collision-aware wireframe printing~\cite{wu2016printing}, stiffness optimization for cable-driven systems~\cite{gueners2020stiffness}, mobile 3D printing with sub-mm accuracy~\cite{li2020slaam}, and extrusion-based fabrication for sensor manufacturing~\cite{munasinghe2021temperature}.
More recent studies have investigated multi-axis and learning-based slicing, including curved slice generation~\cite{zhang2022s3}, representation-agnostic neural slicing~\cite{liu2024neural}, paired CAD--G-code datasets~\cite{jignasu2024slice}, reinforcement learning-based toolpath planning~\cite{huang2024learning, mnih2013playing}, and image-to-G-code generation~\cite{wang2025image2gcode}.
We use Slice-100K~\cite{jignasu2024slice} as our main experimental dataset.
Unlike existing methods, PrintAnything generates printable G-code directly from point clouds, enabling users with 3D-scanned data to obtain fabrication-ready toolpaths.

\input{fig/figure2}

\noindent\textbf{AI for Manufacturing.}
AI has been widely adopted in manufacturing to automate repetitive processes and exploit task-specific data.
Representative directions include shape design~\cite{du2018inversecsg, koch2019abc, yavartanoo2024cnc, yavartanoo2024text2cad}, assembly~\cite{malhan2019identifying, owan2020faster, bhatt2021optimizing, ben2021ikea, schoonbeek2024supervised, wang2025matchmaker}, human monitoring and skill transfer~\cite{kubota2019activity, manyar2023inverse}, visual inspection~\cite{zhang2021defect, yang2023multi, li2025triad}, and factory simulation~\cite{zhang2022smpl}.
Recent studies further address deployment tradeoffs in manufacturing models~\cite{yanglow}, lithography simulation and mask optimization~\cite{zheng2023lithobench}, vision-based automated sewing~\cite{ku2023automated}, reinforcement learning for robotic task sequencing~\cite{manyar2023inverse}, representation learning for assembly state recognition~\cite{schoonbeek2024supervised}, simulation-compatible assembly asset generation~\cite{wang2025matchmaker}, industrial anomaly generation~\cite{tong2025component}, and LLM adaptation for manufacturing inspection~\cite{li2025triad}.

%% file: fig/figure2.tex
\begin{figure*}[t]
\centering
    \includegraphics[width=\linewidth]{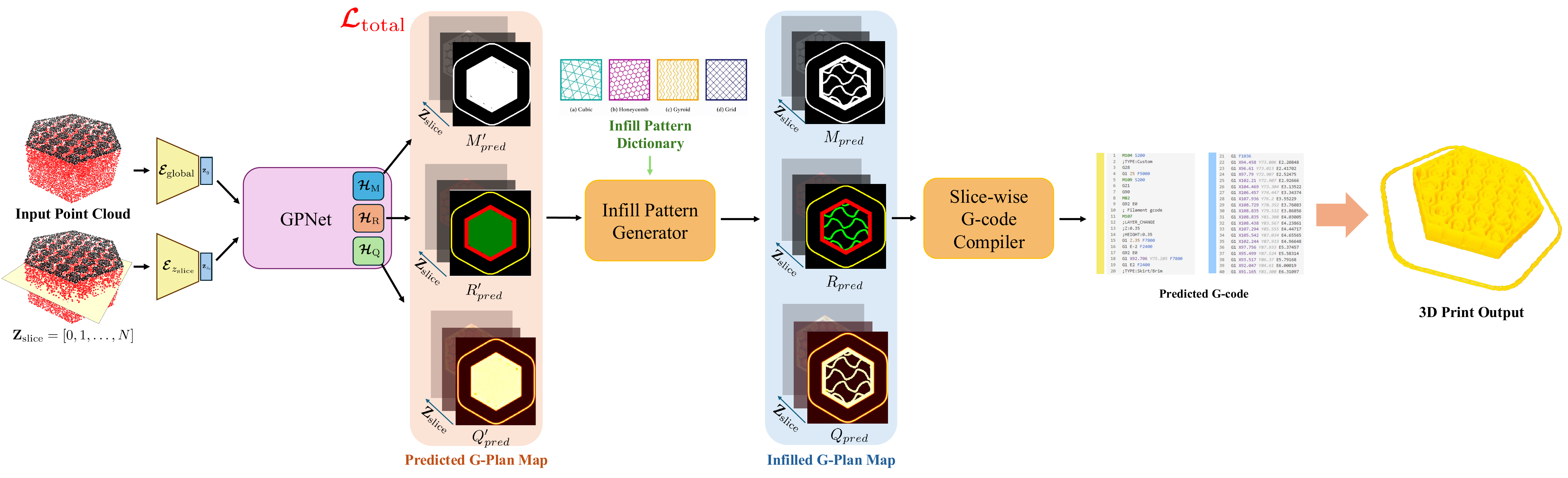}
    \caption{
        \textbf{The framework of PrintAnything.} From input point cloud and slice indices, we encode them and pass to GPNet that predicts M, R, Q. The M, R, Q along with infill pattern dictionary are given to infill pattern generator to generate infilled M, R, Q. Lastly, we feed the infilled M, R, Q to layer-wise G-code generator to get predicted G-code which is directly converted to 3D print output.
    }
    \label{fig:framework}
\end{figure*}

%% file: sec/3_proposed_method.tex
\input{fig/figure3}

\section{Method}
\label{sec:method}

Figure~\ref{fig:framework} illustrates the overall pipeline of PrintAnything, which consists of three stages: geometric plan map prediction, infill pattern generation, and slice-wise G-code generation. 
Prior to training, we construct ground-truth (GT) G-plan maps from the GT G-code and 3D model, which are then used to supervise the training of PrintAnything.

% In this section, we introduce PrintAnything, which is a geometry-conditioned toolpath generation model that predicts printable 3D manufacturing code from 3D shape. Given an input point cloud, we first compute a global shape embedding with a point cloud encoder. We then generate the print layer-by-layer using a high-resolution 2D decoder conditioned on the current layer height and, the previously generated layer to capture inter-layer dependencies. For each layer, the model outputs dense spatial fields including and occupancy mask, region labels, and with flow field. Finally, a deterministic post-processor compiles these predicted fields into standard G-code while producing valid, executable toolpaths.

\subsection{GT Geometric plan (G-plan) map generation}
\label{sec:ir_gt}
Given a pair consisting of a 3D mesh~$\mathcal{M}$ and its corresponding sliced G-code~$\mathbf{G}$, we construct three geometric plan~(G-plan) maps: an occupancy map~$\mathbf{M}$, a region map~$\mathbf{R}$, and a flow map~$\mathbf{Q}$.
The sliced G-code~$\mathbf{G}$ encodes the toolpath coordinates of the 3D printer together with printing attributes, including structural categories (\textit{e.g.,} \texttt{gap}, \texttt{wall}, \texttt{infill}, \texttt{support}, \texttt{skirt}) and the extrusion length along each path segment for a given slice.
For each slice, we first build the region map~$\mathbf{R}$ by rasterizing the structural categories specified in~$\mathbf{G}$.
Based on $\mathbf{R}$, we then construct the occupancy map~$\mathbf{M}$, which represents the union of all printed regions.
Finally, we compute the flow map~$\mathbf{Q}$ by assigning normalized extrusion-length values to the corresponding coordinates.
All maps are resized to a fixed resolution of $256 \times 256$ for compatibility with image-based encoder-decoder frameworks~(\textit{e.g.,} U-Net~\cite{ronneberger2015u}).

% We construct training pairs from the Slice100k~\cite{jignasu2024slice} dataset, which provides a 3D mesh and its corresponding sliced G-code. Since raw G-code is verbose and highly variable across slicer settings, we convert each toolpath into a compact, learning-friendly intermediate representation~(IR) defined on a per-layer 2D grid in the printer's XY plane. For every layer, the IR contains an occupancy map~$\mathbf{M}$ indicating where material is deposited, a region map~$\mathbf{R}$ that assigns each printed location to a coarse semantic category (perimeter, infill, support, skirt, or empty), and a flow map~$\mathbf{Q}$ that encodes the local extrusion intensity. Along with these slice-wise signals, we store the minimal geometric context needed to interpret the grid in metric space. This IR serves as the supervision target for PrintAnything, enabling the model to learn layer-by-layer printing structure without directly modeling the full G-code program.

% \subsection{Training Objective and Optimization}
% We train PrintAnything end-to-end to map an input point cloud to its slice-wise IR targets extracted from Slice100k dataset. Given the global shape embedding and a normalized height $z$, the network predicts the IR for each layer, including an occupancy map $M$, a region map $R$, and a flow map $Q$ can be learned as an auxiliary signal. Trianing uses a muliti-task objectvie: and  

\subsection{Geometric plan map prediction}
Given an input point cloud $\mathbf{P}\in\mathbb{R}^{N\times 3}$, our model predicts G-plan maps defined in Section~\ref{sec:ir_gt}.  
For each slice indexed by $Z_{\mathrm{slice}} \in \{0,1,\dots,N\}$, we estimate an occupancy map~$M_{pred}'$, region map~$R_{pred}'$, and flow map~$Q_{pred}'$.
We denote the normalized slice height as
$\tilde{Z}_{\mathrm{slice}} = \frac{Z_{\mathrm{slice}}}{N} \in [0,1]$.
We first encode the input point cloud using a point-cloud encoder
$\mathcal{E}_{\mathrm{global}}$ (\textit{i.e.,} Point Transformer V3~\cite{wu2024point}), producing a global feature:
$\mathbf{z}_{\mathrm{g}} = \mathcal{E}_{\mathrm{global}}(\mathbf{P}) \in \mathbb{R}^{D}$.
For each slice, we construct a slice-aligned 2D input by projecting the point cloud onto the printer XY grid at height $\tilde{Z}_{\mathrm{slice}}$.
Points satisfying $\lvert z - \tilde{Z}_{\mathrm{slice}}\rvert \le \Delta z/2$ are selected, together with adjacent bins above and below to provide local context which we call Multi-Slice Conditioning~(MSC).
We rasterize these points into an $H\times W$ grid and compute, for each bin, an occupancy mask, a density map, and a height-offset map.
Concatenating these channels yields a tensor:
$\mathbf{H}_{Z_{\mathrm{slice}}} \in \mathbb{R}^{C\times H\times W}$.
The normalized slice height is embedded using an MLP: $\mathbf{z}_{z_{\mathrm{s}}} = \mathcal{E}_{\mathrm{z_{slice}}}(\tilde{Z}_{\mathrm{slice}})$.
Based on global feature~$\mathbf{z}_{\mathrm{g}}$ and embedded slice height~$\mathbf{z}_{z_{\mathrm{s}}}$, we build slice conditioning vector as
$\mathbf{h}_{Z_{\mathrm{slice}}} = [\,\mathbf{z}_{\mathrm{g}};\mathbf{z}_{z_{\mathrm{s}}}\,].$
Then, we use a U-Net~\cite{ronneberger2015u} architecture that consists of an encoder and decoder~(GPNet).
The encoder processes the slice input as:
$\mathbf{X}_{z_{\mathrm{slice}}} =
\mathcal{E}_{\mathrm{GPNet}}(\mathbf{H}_{z_{\mathrm{slice}}})$.
The conditioning vector $\mathbf{h}_{Z_{\mathrm{slice}}}$ is injected into the bottleneck using FiLM~\cite{perez2018film}:
\begin{equation}
\mathbf{X}_{z_{\mathrm{slice}}}' =
\mathbf{X}_{z_{\mathrm{slice}}} \odot (1+\gamma(\mathbf{h}_{z_{\mathrm{slice}}}))
+ \beta(\mathbf{h}_{z_{\mathrm{slice}}}).
\end{equation}
In the end, the decoder produces feature maps:
$\mathbf{Y}_{z_{\mathrm{slice}}} =
\mathcal{D}_{\mathrm{GPNet}}(\mathbf{X}_{z_{\mathrm{slice}}}')$.
Lastly, the final G-plan map are predicted using $1\times1$ convolution heads:
\begin{equation}
M'_{\mathrm{pred}},\;
R'_{\mathrm{pred}},\;
Q'_{\mathrm{pred}}
=
\mathcal{H}_{\mathrm{M}}(\mathbf{Y}_{z_{\mathrm{slice}}}),\;
\mathcal{H}_{\mathrm{R}}(\mathbf{Y}_{z_{\mathrm{slice}}}),\;
\mathcal{H}_{\mathrm{Q}}(\mathbf{Y}_{z_{\mathrm{slice}}}).
\end{equation}
All maps are predicted at resolution $256\times256$.

\subsection{Infill pattern generation}
\label{sec:infill}
Our goal is to choose an infill pattern $p$ and scale $s$ that balance part strength and printing cost. Rather than learning infill geometry from scratch, we leverage a small library of hand-designed templates (e.g., grid, cubic, honeycomb, gyroid) and train a lightweight recommender using automatically generated supervision.
We construct a bootstrap dataset by repeatedly sampling a training shape and randomly selecting a template pattern $p$ and a continuous scale parameter $s$. For each sampled pair $(p,s)$, we warp the corresponding template to the target layer resolution using periodic wrapping so the pattern tiles seamlessly, and insert it only inside the infill-designated regions of the predicted region map $\mathbf{R}$, while keeping structural regions (perimeter/support/skirt) unchanged. This produces a synthesized G-plan map $M_{pred}$, $R_{pred}$, and $Q_{pred}$ for the chosen $(p,s)$. We then compute two inexpensive proxy objectives from the synthesized maps: a strength proxy $S$ and a cost proxy $C$. The strength proxy increases with (i) the infill ratio within occupied regions, (ii) infill connectivity measured by the largest connected-component fraction, and (iii) directional balance; the cost proxy increases with material usage and infill boundary transitions, which serve as a simple proxy for path complexity and printing overhead. Repeating this procedure yields many labeled examples of the form
\begin{equation}
(\phi(M'_{\mathrm{pred}},R'_{\mathrm{pred}}, Q'_{\mathrm{pred}}),\, p,\, s)\ \longrightarrow\ (S,\, C),
\end{equation}
where $\phi(M'_{\mathrm{pred}},R'_{\mathrm{pred}}, Q'_{\mathrm{pred}})$ denotes simple features computed from the predicted G-plan maps.

Using this dataset, we fit a small surrogate recommender with two regressors that predict $\hat{S}(p,s)$ and $\hat{C}(p,s)$ from the candidate choice. At inference time, we enumerate a discrete candidate set $\{(p_i,s_i)\}$, predict $(\hat{S}_i,\hat{C}_i)$ for each candidate, and retain non-dominated solutions that form the Pareto frontier (higher predicted strength with lower predicted cost). We select the final policy from this frontier and generate the final infill by warping and inserting the corresponding template into the predicted G-plan maps.

\subsection{Slice-wise G-code generation}
\label{sec:gcode}
Given the infilled G-plan maps $M_{pred}$, $R_{pred}$, and $Q_{pred}$, we generate G-code by converting each slice into ordered toolpaths on the printer XY plane. 
Each slice provides the pixel-to-millimeter scale $\texttt{px\_mm}$, the slice height, and the raster origin in physical coordinates. 
Before conversion, we center the slice by placing the raster origin at the center of the print bed. 
For a pixel location $\mathbf{u}=(u_x,u_y)$, the physical coordinate at the pixel center is:
\begin{equation}
(x,y) =
\big((u_x + 0.5)\,\texttt{px\_mm} + x_0,\;
     (u_y + 0.5)\,\texttt{px\_mm} + y_0\big),
\end{equation}
where $(x_0,y_0)$ is the centered origin offset in millimeters. We generate perimeter and infill toolpaths from the occupancy mask $M_{pred}$, where the region map $R_{pred}$ determines the structural type of each area (\textit{e.g.,} wall, infill), and compute extrusion for each segment from its physical length, printing parameters, and a flow value sampled from $Q_{pred}$. 
For example, with $\texttt{px\_mm}=0.5$ mm, slice height $\textbf{0.20}$ mm, line width $0.45$ mm, centered origin $(100,100)$ mm, filament diameter $1.75$ mm, and $Q_{pred}=1.0$, three boundary pixels $(20,40)\rightarrow(60,40)\rightarrow(60,80)$ (labeled as \textbf{wall} in $R_{pred}$) map to $(\textbf{110.25},\textbf{120.25})$, $(130.25,120.25)$, and $(130.25,140.25)$ mm, computed by $(20{+}0.5)\!\times\!0.5+100=110.25$ mm and $(40{+}0.5)\!\times\!0.5+100=120.25$ mm for $(20,40)$. 
Each segment has length $20$ mm, yielding a deposited volume of
$0.45 \times 0.20 \times 20 = 1.8\ \text{mm}^3$,
which corresponds to an extrusion length of
$1.8 \div \big(\pi(1.75/2)^2\big) \approx \textbf{0.75}\ \text{mm}$.
Accumulating extrusion along consecutive segments produces G-code such as:
\begin{center}
\begin{tcolorbox}[
  title=Example G-code,
  width=0.62\linewidth,
  boxrule=0.4pt,
  colback=black!3,
  colframe=black!50,
  arc=1mm,
  left=2mm,right=2mm,top=1mm,bottom=1mm,
  listing only
]
\begin{lstlisting}[basicstyle=\ttfamily\small]
; Perimeter (wall)
G1 X110.25 Y120.25 Z0.20 E0.75
G1 X130.25 Y120.25 Z0.20 E1.50
G1 X130.25 Y140.25 Z0.20 E1.05
\end{lstlisting}
\end{tcolorbox}
\end{center}
which move the nozzle along connected segments while depositing material. Processing all slices in increasing height order produces the final slice-wise G-code, where $M_{pred}$ determines printed regions, $R_{pred}$ specifies structural roles, and $Q_{pred}$ controls spatially varying extrusion.

\subsection{Final outputs and loss functions}
During training, the model predicts G-plan maps $M_{pred}'$, $R_{pred}'$, and $Q_{pred}'$ for each slice indexed by $Z_{\mathrm{slice}}$, which are supervised using the ground-truth maps $M$, $R$, and $Q$ described in Section~\ref{sec:ir_gt}.  
We optimize a multi-task objective that combines losses for occupancy, region classification, and flow estimation.
The occupancy loss encourages accurate prediction of printed versus empty regions.  
We use binary cross-entropy (BCE) together with a Dice~\cite{milletari2016v} loss:
\begin{equation}
\mathcal{L}_M =
\mathrm{BCE}(M_{pred}', M)
+
\mathrm{DICE}(M_{pred}', M).
\end{equation}
The region loss supervises the structural labels using weighted cross-entropy:
\begin{equation}
\mathcal{L}_R =
\mathrm{WCE}(R_{pred}', R),
\end{equation}
where WCE denotes weighted cross-entropy.
The flow loss measures the error of the predicted extrusion flow only in printed regions.  
We use a masked Huber~\cite{collins1976robust} loss:
\begin{equation}
\mathcal{L}_Q =
\mathrm{Huber}\!\left(Q_{pred}', Q; M\right),
\end{equation}
where the mask $M$ restricts the loss to printed regions.
The final training objective is the weighted sum of the three losses, where $\lambda_M$, $\lambda_R$, and $\lambda_Q$ are scalar weights that balance the contribution of each term:
\begin{equation}
\mathcal{L}
=
\lambda_M \mathcal{L}_M
+
\lambda_R \mathcal{L}_R
+
\lambda_Q \mathcal{L}_Q.
\end{equation}

%% file: fig/figure3.tex
\begin{figure}[t]
\centering
    \includegraphics[width=1.0\linewidth]{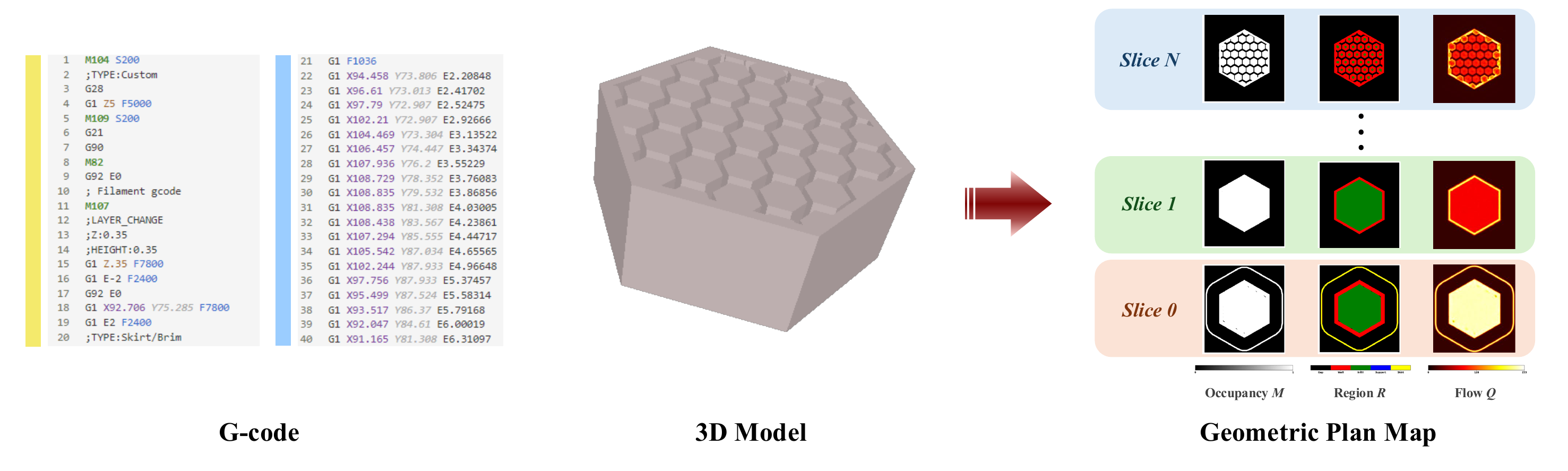}
    \caption{
        \textbf{GT G-plan map generation.} We generate geometric plan~(G-plan) map from a pair of G-code and 3D model.
    }
    \label{fig:gt_generation}
\end{figure}

%% file: sec/4_implementation_details.tex
\section{Implementation details}
\label{sec:implementation_details}
We implement PrintAnything in PyTorch~\cite{paszkepytorch} and train with AdamW~\cite{loshchilov2018decoupled} using mixed precision.
We sample $N{=}30{,}000$ surface points per shape and normalize them to $[-1,1]$.
All G-plan maps are rasterized and predicted at a fixed resolution of $256{\times}256$ per slice.
Following Section~\ref{sec:method}, we encode the input point cloud with a global point-cloud encoder $E_{\text{global}}$ (Point Transformer V3~\cite{wu2024point}), producing a global feature $\mathbf{z}_g$.
For each slice, we build a slice-aligned 2D input by projecting points near the target height including adjacent bins above and below for local context onto the printer XY grid.
We embed the normalized slice height $\tilde{z}\in[0,1]$ with an MLP and concatenate it with $\mathbf{z}_g$ to form the slice conditioning vector.
We use a U-Net~\cite{ronneberger2015u} backbone for GPNet to process the slice input, inject the conditioning at the bottleneck via FiLM~\cite{perez2018film}, and predict occupancy, region, and flow maps using separate $1{\times}1$ convolution heads. We train for 50 epochs with learning rate $2{\times}10^{-4}$ and weight decay $10^{-4}$. All experiments are conducted on a single NVIDIA Quadro RTX 8000 GPU.
For all generated G-code, we use PrusaSlicer~\cite{prusaslicer} to simulate the printing process and visualize the 3D prints.

%% file: sec/5_experiments.tex
\section{Experiments}
\label{sec:experiments}

\subsection{Datasets}
% TODO: WRITE THIS
We conduct our experiments on Slice-100K~\cite{jignasu2024slice}, a large-scale dataset for learning printing-oriented 3D representations. Each sample contains a 3D shape together with manufacturing supervision, including its CAD model (STL) and the corresponding ground-truth G-code produced by a standard slicing pipeline. We randomly partition the shapes into training and test sets with a 9:1 ratio and report mean performance on the held-out test set.

As input, we represent each shape as a surface point cloud: we uniformly sample 30k points and normalize coordinates to the range $[-1,1]$. For supervision and model outputs, we construct layer-wise G-plan maps by rasterizing the ground-truth G-code into occupancy, region, and flow maps at the target slice resolution, which are used to train and evaluate our model.

\subsection{Evaluation Metrics}
\label{sec:eval-metrics}
In all experiments, we report the average performance over the test set of Slice-100K~\cite{jignasu2024slice} dataset.
For evaluating 3D geometry, our main evaluation metric is Chamfer Distance~(CD) and F1-score~($F1_{3D}$).
Let $\mathcal{P}$ denote points sampled from the predicted shape and $\mathcal{G}$ denote points sampled from the ground-truth (GT) mesh.
We evaluate with Chamfer Distance~(CD) defined as :
\begin{equation}
\mathrm{CD}(\mathcal{P},\mathcal{G})=
\frac{1}{|\mathcal{P}|}\sum_{p\in\mathcal{P}}\min_{g\in\mathcal{G}}\|p-g\|_2
+
\frac{1}{|\mathcal{G}|}\sum_{g\in\mathcal{G}}\min_{p\in\mathcal{P}}\|g-p\|_2 .
\end{equation}
We also report F1 score for 3D points after setting a distance threshold~$\tau=1$\,mm:
\begin{equation}
F1_{3D}=\frac{2PR}{P+R},
\end{equation}
where $P$ and $R$ denote precision and recall computed between $\mathcal{P}$ and $\mathcal{G}$.
To evaluate slice-wise structural accuracy, we rasterize slices from the GT G-code and compute a binary occupancy F1 score per each rasterized slices:
\begin{equation}
F1_{2D}=\frac{2\,\mathrm{P}_{2D}\,\mathrm{R}_{2D}}{\mathrm{P}_{2D}+\mathrm{R}_{2D}},
\end{equation}
where $\mathrm{P}_{2D}$ and $\mathrm{R}_{2D}$ denote precision and recall of predicted slice occupancy.
The final score is obtained by averaging across slices and samples.
For the ablation studies on G-plan map design, we further evaluate extrusion-flow properties derived from the generated G-code.
We consider only extruding moves ($\Delta E_k>0$) with non-zero XY motion.
For each move, we define the planar travel length
$\Delta s_k = \|(x_k,y_k)-(x_{k-1},y_{k-1})\|_2$ and the extrusion density
$\rho_k=\frac{\Delta E_k}{\Delta s_k+\epsilon}$.
We measure extrusion smoothness using:
\begin{equation}
\Delta\rho\text{-smooth}=
\frac{1}{N-1}\sum_{k=2}^{N}\left|\rho_k-\rho_{k-1}\right|,
\end{equation}
which captures local variation of extrusion density along toolpaths.
We also evaluate per-slice extrusion consistency by computing the coefficient of variation of extrusion density within each slice:
\begin{equation}
\mathrm{CV}_z=
\frac{\mathrm{std}(\rho^{(z)})}{|\mathrm{mean}(\rho^{(z)})|+\epsilon},
\qquad
\rho\text{-CV}=
\frac{1}{Z'}\sum_{z=1}^{Z'} \mathrm{CV}_z ,
\end{equation}
where $\rho^{(z)}$ denotes the extrusion densities of moves within slices $z$, and $Z'$ is the number of slices containing at least one extruding move.
Lower values indicate smoother and more consistent extrusion behavior.

\input{fig/figure_qual}
\input{tab/sota_comparison}

\subsection{Comparison with state-of-the-art methods}
\label{sec:sota}

As noted in section~\ref{sec:intro}, a natural and widely adopted baseline for printing from point observations is to first convert the input into a mesh and then rely on a mature slicer to produce G-code. Following this principle, we compare against strong mesh-conversion pipelines built on classical Poisson surface reconstruction~\cite{kazhdan2006poisson} and recent learning-based methods, DWG~\cite{liu2025diffusing} and MeshAnything~\cite{chenmeshanything}. 
For each baseline, we reconstruct a mesh from the input point cloud and feed the resulting mesh into PrusaSlicer~\cite{prusaslicer} to generate the final toolpath, yielding a standard ``point cloud $\rightarrow$ mesh $\rightarrow$ slicer $\rightarrow$ G-code'' workflow. 

Table~\ref{tab:sota_contrastive} summarizes the results on Slice-100K~\cite{jignasu2024slice}.
Our method achieves the best performance across all metrics, reaching $0.047$ CD, $0.7414$ $F1_{3D}$, and $0.677$ $F1_{2D}$.
In comparison, mesh-based pipelines tend to accumulate errors across stages: small geometric artifacts introduced during reconstruction can be amplified after slicing, which is reflected most clearly in the lower slice-level scores. 
We further visualize output 3D prints in Figure~\ref{fig:qual}.
Poisson reconstruction often smooths away thin structures, while learning-based mesh methods may introduce broken surfaces or local clutter, which can translate into missing or fragmented toolpaths after slicing. In contrast, our approach produces more coherent and complete print structures, particularly on shapes with slender parts and sharp features, leading to toolpaths that better match the target slices. Overall, these quantitative and qualitative results indicate that representing the input with our G-plan map yields more reliable, slice-faithful toolpaths than mesh-based conversion pipelines followed by a conventional slicer.

% \subsection{Continuous/ arbitrary z prediction}
% qualitative

\input{tab/multi_layer_conditioning}

\subsection{Multi-slice conditioning}
\label{subsec:mlc}
Table~\ref{tab:abl_mlc} shows the impact of multi-slice conditioning~(MSC) on Slice-100K~\cite{jignasu2024slice}.
Enabling MSC notably improves geometric accuracy: CD drops from 0.059 to 0.047, a 20.3\% reduction.
It also improves 3D shape fidelity, with $\mathrm{F1}_{\mathrm{3D}}$ increasing from 0.702 to 0.741, which is a 5.6\% improvement.
These gains suggest that conditioning on multiple neighboring slices helps the model resolve slice-wise ambiguities.
By enforcing inter-slice continuity and structural support cues, our PrintAnything leads to more consistent toolpath decisions than relying only on a single-slice context.

\input{tab/infill}

\subsection{Infill policy recommendation}
\label{subsec:recommender}
Table~\ref{tab:recommender_vs_random_baselines} compares our infill policy recommender with random-scale and fixed-scale baselines across a diverse set of infill patterns.
We evaluate each method using three criteria: a proxy for structural strength, a proxy for fabrication cost, and a combined score that balances the two.
Concretely, the strength proxy measures (i) the infill material ratio within occupied regions, (ii) the connectivity of the infill (largest connected-component ratio), and (iii) a direction-balance term that favors well-distributed infill boundaries, while the cost proxy increases with total occupied area and the number of infill boundary transitions. Detailed computation of these measures is included in the Supplementary Details.
We compute the combined score as $\text{Combined}=10^{5}\cdot \text{Strength}/\text{Cost}$.

Among the baselines, random-scale strategies can occasionally achieve competitive strength.
For example, Grid + Random Scale attains the highest strength value of 0.310, but it incurs a high cost of 41,535 and therefore yields a low combined score of 0.748.
In contrast, our recommender achieves a strong strength value of 0.308 while producing the lowest fabrication cost of 32,974.
As a result, it obtains the best combined score of 0.937.
Overall, these results indicate that our recommender learns to select infill strategies that better balance the strength--cost trade-off, leading to more efficient yet reliable fabrication outcomes.

\input{fig/figure_qual_raw}

\subsection{Robustness to scan artifacts}
\label{subsec:robustness_scan}
In practical settings, point clouds acquired from commodity depth sensors or 3D scanners are often affected by occlusions and measurement noise.
To encourage robustness under such scan artifacts, we additionally apply two augmentations during training: structured hole dropout, which removes points inside a contiguous 3D spherical region to mimic occlusion, and additive Gaussian noise with standard deviation $0.01$ in normalized coordinates. 

Figure~\ref{fig:qual_rawa} shows qualitative results for incomplete point clouds with a missing circular portion of the geometry, a common failure mode caused by sensor-object occlusion.
Despite the missing observations, our method recovers a complete printable shape, producing stable occupancy predictions and coherent toolpaths. Figure~\ref{fig:qual_rawa} presents a noisy input case. Even when the point cloud is corrupted by noise, the projected slices remain stable, and both the predicted occupancy map and the generated G-code preserve the intended structure without introducing spurious artifacts. Overall, these results indicate that our pipeline is robust to realistic scan imperfections and can reliably generate printable toolpaths from imperfect point observations.
% real-world noisy, incomplete

\input{tab/abl_ir_design}

\subsection{Ablations on G-plan maps design}
\label{subsec:abl_ir_design}
We analyze the design of the proposed geometric plan~(G-plan) map by incrementally adding its components: $\mathbf{M}$, $\mathbf{R}$, $\mathbf{Q}$.
Specifically, we compare three configurations: using only the occupancy map~$\mathbf{M}$~(first row), using occupancy map~$\mathbf{M}$ and region map~$\mathbf{R}$~(second row), and using the full representation with the flow map~$\mathbf{Q}$~(third row).
Table~\ref{tab:abl_ir_design} reports the results on Slice-100K~\cite{jignasu2024slice}.
Using only $\mathbf{M}$ provides a coarse printable geometry, but it cannot distinguish between structurally different regions that require different deposition behaviors.
Region map~$\mathbf{R}$ improves geometric fidelity, reducing CD by 11.5\% and increasing $\mathrm{F1}_{\mathrm{3D}}$ by 2.7\%.
This improvement suggests that region-level cues help disambiguate toolpath assignment around boundaries and interior supports.
Adding the flow map~$\mathbf{Q}$ further improves flow-related metrics of the predicted toolpaths.
% While the geometric metrics remain comparable, the flow-related metrics improve significantly.
In particular, the flow smoothness measure $\Delta\rho$-smooth decreases from $0.081$ to 0.035, corresponding to a 56.8\% reduction, and the extrusion variability $\rho$-CV decreases from $1.059$ to 0.909, which is a 14.2\% reduction.
% These improvements indicate that $\mathbf{Q}$ helps the model maintain more consistent extrusion rates along toolpaths, leading to smoother material deposition and more stable printing behavior.
Overall, the results highlight that all components of our G-plan map are critical not only for geometric reconstruction but also for achieving extrusion consistency per path length during G-code generation.

\input{fig/Real_machine_output}

\subsection{Results on real-world 3D printer machines}
\label{sec:supp_real_3d_print}

Figure~\ref{fig:real_machine_output} presents real-world 3D printing results produced directly from the G-code generated by PrintAnything. 
All objects were fabricated using a Bambu Lab X1-Carbon printer without any additional post-processing of the generated G-code, allowing us to evaluate the raw printing quality of our method. 
The results demonstrate that the predicted G-code is robust across a variety of shapes and geometric complexities. 
In particular, the first, second, third, and seventh rows show mechanical parts with cogwheel structures commonly used in manufacturing, where the printed outputs exhibit consistent and well-formed teeth across the entire gear. 
The fourth row illustrates a complex object with thin and delicate structures, for which PrintAnything still produces detailed and stable prints. 
We further evaluated the structural integrity of the printed objects through manual pressure tests, where the printed parts remained intact under reasonable external force. 
These results indicate that PrintAnything can generate reliable G-code that translates effectively into high-quality physical prints on real-world 3D printing hardware.

% Flow Smoothness

% For Q validation
% -> Extrusion Consistency per Path Length

% \subsection{Efficiency}
% speed
% toolpath (more important)

% \subsection{Real-World Validation}
% Shows results on real-world simulator/printer

% % qualitative

% % text-to-point cloud
% % image-to-point cloud

%% file: fig/figure_qual.tex
\begin{figure*}[t]
\centering
    \includegraphics[width=\linewidth]{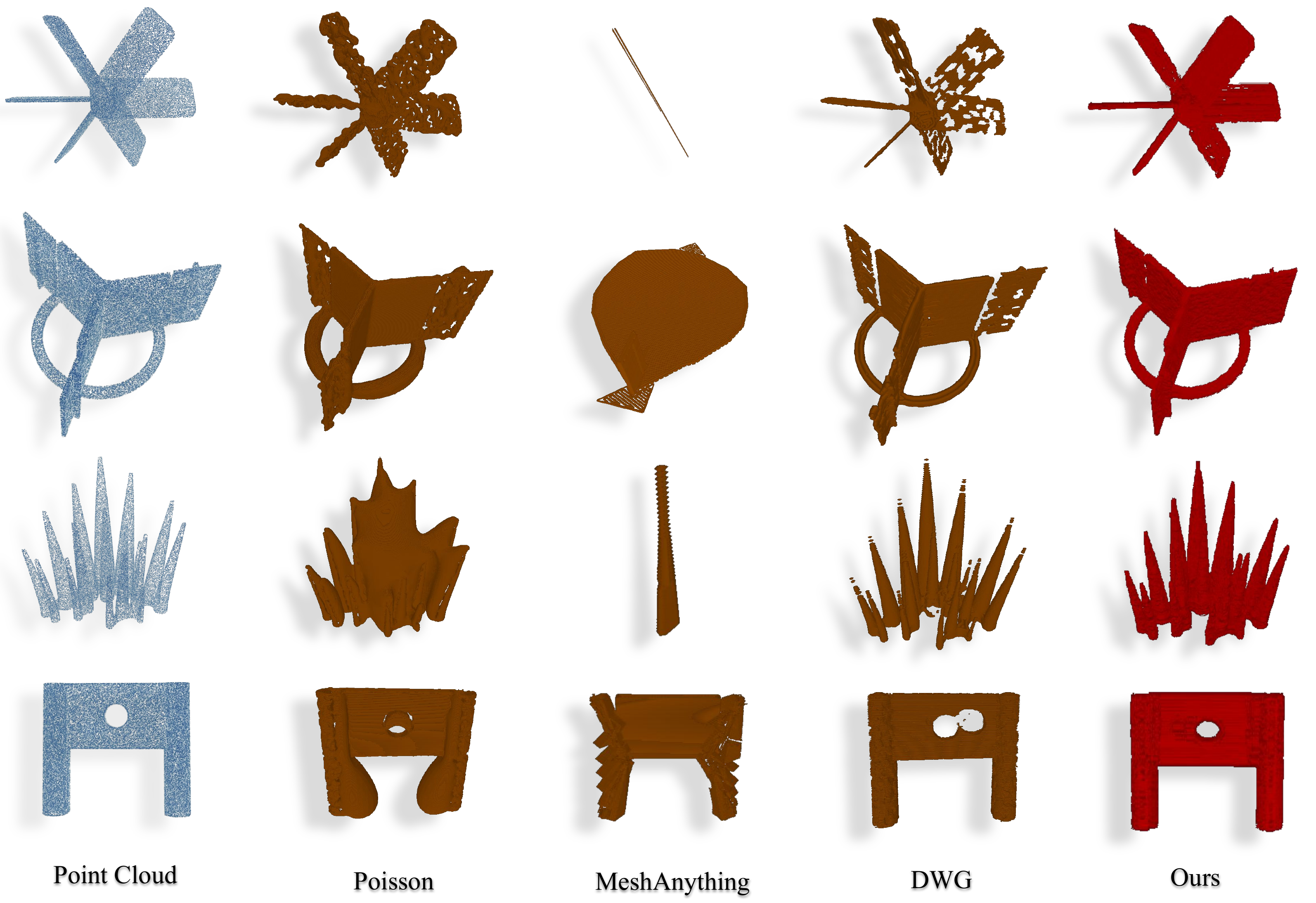}
    \caption{
        \textbf{Qualitative Results on Slice-100K~\cite{jignasu2024slice} dataset.}
    }
    \label{fig:qual}
\end{figure*}

%% file: tab/sota_comparison.tex
\begin{table}[htbp]
\centering
\footnotesize
\setlength{\tabcolsep}{2pt}
\caption{\textbf{Quantitative comparison of various point-to-G-code prediction techniques on Slice-100K~\cite{jignasu2024slice} dataset.}}
\begin{tabular}{@{}llccc@{}}
\toprule
Representation & Method & CD~$\downarrow$ & $\text{F1}_{\text{3D}}{\uparrow}$ & $\text{F1}_{\text{2D}}{\uparrow}$ \\
%& Infill Acc.~$\uparrow$ \\ 
\midrule

\multirow{3}{*}{Mesh} 
& Poisson~\cite{kazhdan2006poisson} & 0.088 & 0.682 & 0.587  \\
% & Point2Mesh~\cite{hanocka2020point2mesh} &  &  &  &  \\
% & Shape as points~\cite{peng2021shape} &  &  &  &  \\
%& iPSR~\cite{hou2022iterative} &  &  &  &  \\
& DWG~\cite{liu2025diffusing} & 0.062 & 0.712 & 0.496  \\
& MeshAnything~\cite{chenmeshanything} & 0.157 & 0.480 & 0.356  \\
\midrule

% \multirow{2}{*}{Implicit-based}
% & Points2Surf~\cite{erler2020points2surf} &  &  &  & 
% \\
% & POCO~\cite{boulch2022poco} &  &  &  & 
% \\

% \midrule

% \multirow{2}{*}{Voxel-based}
% & PointToVoxel~\cite{spconv2022} &  &  &  &  \\
% & Open3D~\cite{zhou2018open3d} &  &  &  &  \\

% \midrule

% Wireframe-based 
% & Huang~\textit{et al.}~\cite{huang2024learning} &  &  &  &  \\

% \midrule

% Image-based 
% & Image2Gcode~\cite{wang2025image2gcode} &  &  &  &  \\

G-plan map
& PrintAnything~(Ours) & \textbf{0.047}  & \textbf{0.741} & \textbf{0.677} \\

\bottomrule
\end{tabular}
\label{tab:sota_contrastive}
\end{table}

%% file: tab/multi_layer_conditioning.tex
\begin{table}[htbp]
\centering
\footnotesize
\setlength{\tabcolsep}{5pt}
\caption{\textbf{Ablation of multi-slice conditioning (MSC) on Slice-100K~\cite{jignasu2024slice}.}}
\begin{tabular}{@{}cccc@{}}
\toprule
Multi-slice conditioning & CD~$\downarrow$ & $\mathrm{F1}_{\mathrm{3D}}~\uparrow$ & $\mathrm{F1}_{\mathrm{2D}}~\uparrow$ \\
\midrule
\ding{55} & 0.059  & 0.702 & 0.652  \\
\ding{51} & \textbf{0.047} & \textbf{0.741} & \textbf{0.677} \\
\bottomrule
\end{tabular}
\vspace{-8mm}
\label{tab:abl_mlc}
\end{table}

%% file: tab/infill.tex
\begin{table}[htbp]
\centering
\small
\setlength{\tabcolsep}{4pt}
\renewcommand{\arraystretch}{1.1}
\caption{\textbf{Comparison of our recommender against random-scale/fixed-scale baselines on Slice-100K~\cite{jignasu2024slice}.}}
\begin{tabular}{p{0.45\linewidth}ccc}
\toprule
Method & Strength $\uparrow$ & Cost $\downarrow$ & Combined$\uparrow$ \\
\midrule
Random Pattern + Random Scale & 0.294 & 38,808 & 0.757 \\
Cubic + Random Scale & 0.298 & 36,994 & 0.806 \\
Grid + Random Scale & \textbf{0.310} & 41,535 & 0.748 \\
Gyroid + Random Scale & 0.271 & 42,247 & 0.643 \\
Honeycomb + Random Scale & 0.294 & \underline{35,445} & 0.830 \\
Random Pattern + Fixed Scale ($s{=}1.0$) & 0.300 & 35,866 & \underline{0.837} \\
\textbf{Ours (Recommender)} & \underline{0.308} & \textbf{32,974} & \textbf{0.937} \\
\bottomrule
\end{tabular}
\label{tab:recommender_vs_random_baselines}
\vspace{-3mm}
\end{table}

%% file: fig/figure_qual_raw.tex
\begin{figure}[t]
    \centering
    
    \includegraphics[width=\linewidth]{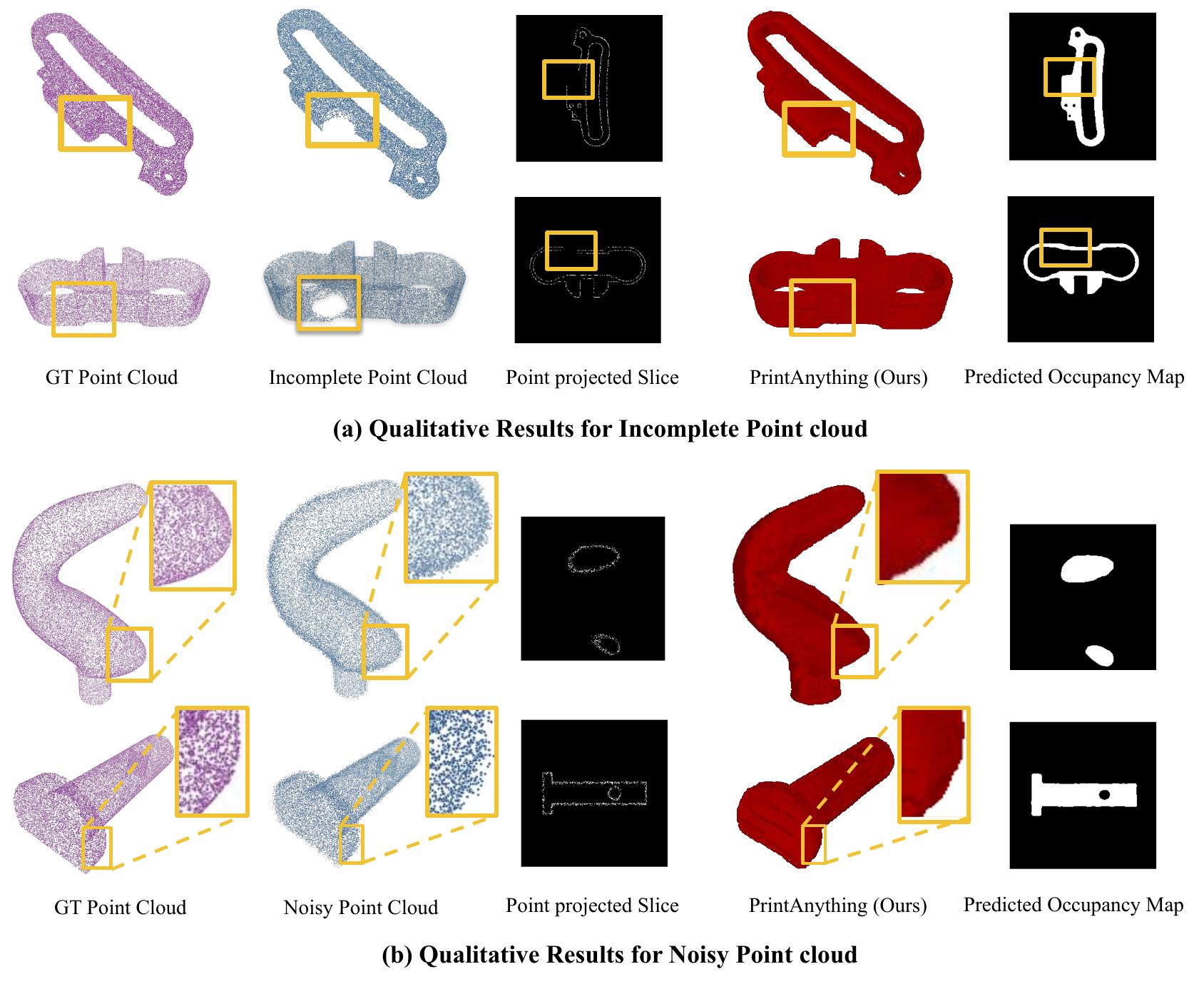}
    
    % This line forces the next caption to ignore the "a" and "b" count
    \setcounter{subfigure}{0} 
    
    \caption{
        \textbf{Qualitative Results on incomplete and noisy point cloud input on Slice-100K~\cite{jignasu2024slice}. Yellow boxes highlight regions with missing observations for incomplete input. For the noisy case, enlarged insets are provided for clearer comparison.}
    }
    \vspace{-4mm}
    \label{fig:qual_rawa}
\end{figure}

%% file: tab/abl_ir_design.tex
\begin{table}[hbtp]
\centering
\footnotesize
\setlength{\tabcolsep}{2.7pt}
\vspace{-2mm}
\caption{\textbf{Ablation of representations on Slice-100K~\cite{jignasu2024slice}.}}
\begin{tabular}{@{}cccccccc@{}}
\toprule
M & R & Q & CD~$\downarrow$ & $\mathrm{F1}_{\mathrm{3D}}\uparrow$ & $\mathrm{F1}_{\mathrm{2D}}\uparrow$ &
$\Delta\rho$-smooth~$\downarrow$ & $\rho\text{-CV}~\downarrow$ \\
\midrule
\ding{51} & \ding{55} & \ding{55} & 0.052 & 0.709 & 0.635 & 0.082 & \underline{1.055} \\
\ding{51} & \ding{51} & \ding{55} & \textbf{0.046} & \underline{0.728} & \underline{0.675} & \underline{0.081} & 1.059 \\
\ding{51} & \ding{51} & \ding{51} & \underline{0.047} & \textbf{0.741} & \textbf{0.677} & \textbf{0.035} & \textbf{0.909} \\
\bottomrule
\end{tabular}
\label{tab:abl_ir_design}
\vspace{-2mm}
\end{table}

%% file: fig/Real_machine_output.tex
\begin{figure*}[p]
\centering
    \includegraphics[width=\linewidth]{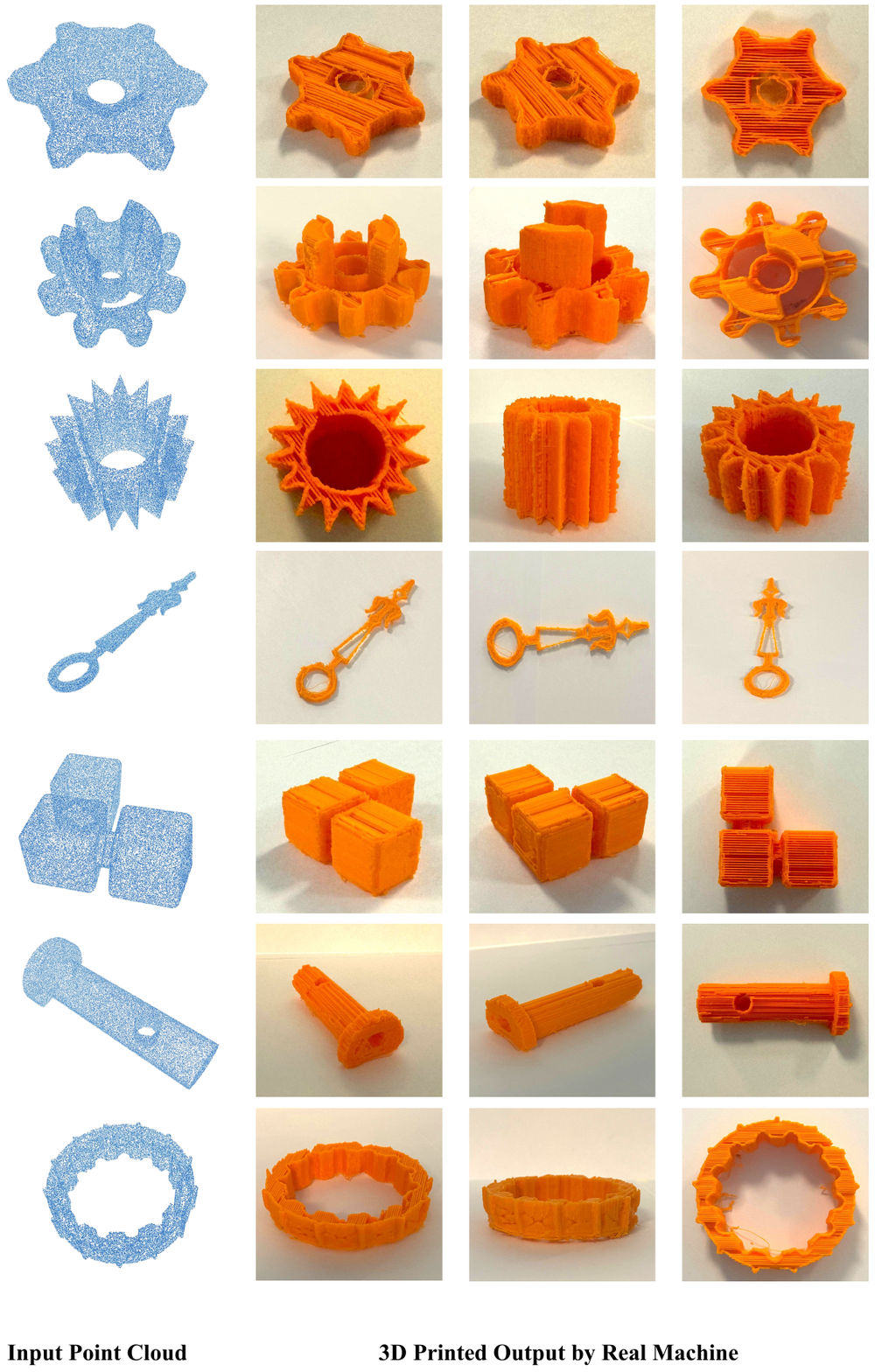}
    \caption{
    \textbf{Visualization of 3D printed output by Real Machine.}
    }
    \label{fig:real_machine_output}
\end{figure*}

%% file: sec/6_conclusion.tex
\section{Conclusion}
\label{sec:conclusion}
We present PrintAnything, a framework that generates executable 3D printing G-code directly from point clouds without requiring mesh reconstruction. 
To bridge point clouds and slice-wise fabrication, we introduce a slice-wise point projection strategy that converts point clouds into slice-aligned 2D representations. 
We further propose the Geometric plan (G-plan) map, composed of occupancy, region, and flow maps that encode the geometric and extrusion properties required for toolpath synthesis. 
In the end, PrintAnything predicts G-plan maps from the input point cloud and converts them into executable toolpaths to produce printable G-code, enabling a practical mesh-free fabrication pipeline. As future work, we plan to extend the framework to support more diverse printer settings and materials with physical constraints. 

%% file: sec/acknowledgement.tex
\noindent
\section*{Acknowledgement}
This work was supported in part by the IITP grants [No. RS-2021-II211343, Artificial Intelligence Graduate School Program (Seoul National University), No. RS-2024-00426853, No. RS-2025-02303870, No.2022-0-00156] funded by the Korea government (MSIT).
We thank Juhyoung Lee for assistance with the physical 3D printing and sample fabrication.

%% file: sec/X_suppl.tex
\begin{center}
\textbf{\large Supplementary Material \textit{for} \\ \vspace{2mm}
\large{``PrintAnything: Learning Geometric Plan Map \\for 3D Printing G-code Generation \\from Unoriented Point Clouds"}}
\end{center}

\setcounter{section}{0}
\setcounter{table}{0}
\setcounter{figure}{0}
\renewcommand{\thesection}{S\arabic{section}}
\renewcommand{\thetable}{S\arabic{table}}   
\renewcommand{\thefigure}{S\arabic{figure}}

This supplementary material includes additional experiments, analyses, and results that could not be presented in the main manuscript due to space constraints.
The contents are summarized below:
\begin{compactitem}
    \item \ref{sec:supp_gcode_compiler}. Details of slice-wise G-code compiler
    % \item \ref{sec:supp_gcode_out}. Details of G-code output
    \item \ref{sec:supp_gplan_map_gen}. Details of GT G-plan map generation
    \item \ref{sec:supp_model_arch}. Details of model architecture
    \item \ref{sec:supp_loss_func}. Details of loss functions
    % \item \ref{sec:supp_sota_comp}. Additional comparisons with state-of-the-art methods
    \item \ref{sec:supp_metrics}. Detailed computation of strength and cost proxies
    \item \ref{sec:supp_more_qual_unseen}. Qualitative results on unseen data
    \item \ref{sec:supp_vis_gplan_map}. Visualization of predicted G-plan map
    \item \ref{sec:supp_vis_recomm_gplan_map}. Visualization of infilled G-plan map
    % \item \ref{sec:supp_more_incomplete_noisy}. Additional results on incomplete and noisy point cloud
    % \item \ref{sec:supp_more_qual}. Additional qualitative results
    \item \ref{sec:supp_error_mesh}. Results of erroneous mesh reconstruction
    % \item \ref{sec:supp_real_3d_print}. Results of real-world 3D printer machines
    \item \ref{sec:supp_efficiency}. Running time comparison
    \item \ref{sec:supp_limit_and_societal}. Limitations and societal impacts
\end{compactitem}

\section{Details of slice-wise G-code compiler}
\label{sec:supp_gcode_compiler}

The Algorithm~\ref{alg:gplan_to_gcode} describes the slice-wise procedure that converts the predicted G-plan maps into executable G-code for 3D printing. The compiler takes as input three layer-wise maps---the occupancy mask \(M\), region map \(R\), and flow map \(Q\)---together with printer parameters such as line width \(w\), layer height \(h\), number of perimeter walls \(K\), infill density \(\rho\), filament diameter \(d_{\mathrm{fil}}\), and printing speeds. For each layer, the algorithm first extracts contours from the occupancy mask to generate perimeter paths. Starting from the outer boundary, each contour is simplified, converted into printer coordinates, and assigned region-specific attributes using \(R\). The extrusion amount for a contour segment of length \(\ell\) is computed from the base extrusion \(E_{\mathrm{base}} = \frac{w h \ell}{A_f}\), where \(A_f = \pi(d_{\mathrm{fil}}/2)^2\) is the filament cross-sectional area, and is then modulated by the flow value sampled from \(Q\) at the segment midpoint. After one perimeter wall is completed, the working mask is eroded to generate the next inner wall, and this process is repeated for \(K\) walls. The algorithm then generates infill within the occupied region. Scanline spacing is determined by the target infill density \(\rho\), and occupied intervals are identified from the occupancy mask. For each interval of length \(\ell\), the extrusion is computed using the same base formulation and modulated by the sampled flow value from \(Q\). The resulting perimeter and infill moves are sequentially written into a standard G-code file together with printer-specific headers and footers, yielding executable instructions for fused filament fabrication~(FFF) printers.

\include{algo/slice-wise-conversion}

% \section{Details of G-code output}
% \label{sec:supp_gcode_out}

\section{Details of GT G-plan map generation}
\label{sec:supp_gplan_map_gen}
Algorithm~\ref{alg:gcode_to_gplan} describes the procedure used to generate ground-truth G-plan maps from G-code files. 
The input is a G-code file together with optional rasterization parameters such as pixel size $p$, filament diameter $d_f$, and nozzle width $d_n$. 
The algorithm first parses the G-code file to extract deposited toolpath segments. 
For each line in the file, the parser updates the printer state and checks whether the command corresponds to an extrusion motion. 
If so, the command is converted into a deposited segment containing its endpoints, layer height, extrusion width, printing speed, and motion type, and the segment is appended to the corresponding layer. 
After parsing all toolpaths, the algorithm estimates printing parameters from the G-code header if available, including filament diameter and nozzle width. 
If the raster pixel size is not explicitly specified, it is determined based on the nozzle width to maintain an appropriate spatial resolution.
Next, the algorithm computes a shared raster grid for all layers by collecting the endpoints of all deposited segments and determining a global 2D bounding box. 
Each layer is then rasterized into three maps that constitute the G-plan representation: the occupancy mask $M$, region map $R$, and flow map $Q$. 
For every deposited segment in the layer, the algorithm determines its extrusion width, layer height, region label, and volumetric flow, and rasterizes the segment onto the grid according to its local width. 
For each covered pixel, the occupancy mask $M$ is marked as occupied, the region map $R$ is assigned or updated with the corresponding region label, and the flow map $Q$ accumulates the flow contribution of the segment. 
After processing all segments in the layer, the accumulated flow values are normalized over the covered pixels to obtain the final flow map. 
The resulting $(M,R,Q)$ maps are saved for each layer together with a manifest file that records raster metadata, printer and material parameters, and the file paths of the generated maps.

\input{algo/gcode-to-gplan1}
\input{algo/gcode-to-gplan2}

\section{Details of model architecture}
\label{sec:supp_model_arch}

Our model consists of a 3D point-cloud encoder and a 2D layer decoder. The encoder extracts a global shape feature from the input point cloud, and the decoder predicts the layer-wise G-plan maps $(M,R,Q)$ for each queried printing height.

\noindent\textbf{PTv3 encoder.}
We use Point Transformer V3~(PTv3)~\cite{wu2024point} to encode the input point cloud $\mathbf{P}\in\mathbb{R}^{1\times N\times C}$, where the first three channels are normalized 3D coordinates. We set the voxel size to $0.06$ and use PTv3 in classification mode, i.e., only the embedding stem and encoder are used. The encoder has five stages with channel dimensions $(32,64,128,256,512)$, depths $(2,2,2,2,2)$, attention heads $(2,4,8,16,32)$, and patch size $128$ throughout. PTv3 outputs per-point features of dimension $512$, which are globally max-pooled and linearly projected to obtain a global feature vector $\mathbf{g}\in\mathbb{R}^{1\times512}$.

\noindent\textbf{Layer-wise decoder.}
Given the global feature $\mathbf{g}$, we decode the G-plan map one layer at a time using a 2D U-Net~\cite{ronneberger2015u}. For each queried layer height $z$, the decoder takes as input: 1) the previous-layer occupancy map, and 2) point-cloud hint rasters around the queried height. Each hint slab contains three channels: occupancy, density, and mean height offset. Using three slabs $\{-1,0,+1\}$ gives $9$ hint channels, so the total decoder input has $10$ channels.

The queried height $z\in[0,1]$ is embedded by an MLP and fused with the global feature $\mathbf{g}$. This conditioning is injected into the U-Net bottleneck using FiLM~\cite{perez2018film}:
\[
\mathbf{x}_{\mathrm{bot}} \leftarrow \mathbf{x}_{\mathrm{bot}}\odot(1+\gamma)+\beta.
\]

\noindent\textbf{U-Net and output heads.}
The decoder uses a standard U-Net with base width $32$, encoder channels
\[
10 \rightarrow 32 \rightarrow 64 \rightarrow 128 \rightarrow 256 \rightarrow 512,
\]
and a symmetric decoder with skip connections and bilinear upsampling. The final shared feature map has 64 channels. From this map, we predict the region map $R$ with a $1\times1$ convolution over five classes (\emph{gap}, \emph{wall}, \emph{infill}, \emph{support}, and \emph{skirt}), and the flow map $Q$ with another $1\times1$ convolution. The occupancy mask $M$ is derived from the region prediction by treating all non-gap classes as occupied:
\[
p(M=1)=1-\mathrm{softmax}(R)_{\text{gap}}.
\]

% Overall, the PTv3 encoder captures the global 3D structure of the shape, while the FiLM-conditioned 2D U-Net predicts high-resolution layer-wise G-plan maps.

\section{Details of loss functions}
\label{sec:supp_loss_func}

During training, the model predicts G-plan maps $M'_{\mathrm{pred}}$, $R'_{\mathrm{pred}}$, and $Q'_{\mathrm{pred}}$ for each slice indexed by $Z_{\text{slice}}$. These predictions are supervised using the corresponding ground-truth maps $M$, $R$, and $Q$. We optimize the following multi-task objective:
\[
L=\lambda_M L_M+\lambda_R L_R+\lambda_Q L_Q,
\]
where $\lambda_M=1$, $\lambda_R=1$, and $\lambda_Q=3\times 10^{-5}$ in our implementation.

To reduce memory and computation, we do not supervise all slices in every iteration. Instead, for each training sample, we always include the top and bottom slices and additionally sample up to 20 intermediate slices at random. Let $S$ denote the selected slice set. The final training loss is averaged over the selected slices:
\[
L_{\mathrm{iter}}=\frac{1}{|S|}\sum_{Z_{\text{slice}}\in S}
\left(
\lambda_M L_M+\lambda_R L_R+\lambda_Q L_Q
\right).
\]

The occupancy loss supervises printed versus empty regions. Following the main manuscript, we combine binary cross-entropy~(BCE) and Dice~\cite{milletari2016v} loss:
\[
L_M=\mathrm{BCE}(M'_{\mathrm{pred}},M)+\mathrm{DICE}(M'_{\mathrm{pred}},M).
\]
In implementation, BCE is applied on logits, while the Dice term is computed from the sigmoid probabilities. Concretely, letting
\[
P_M=\sigma(M'_{\mathrm{pred}}),
\]
the Dice term is implemented as:
\[
\mathrm{DICE}(M'_{\mathrm{pred}},M)
=
1-\frac{2\sum(P_M\cdot M)+\epsilon}{\sum P_M+\sum M+\epsilon},
\]
where $\epsilon=10^{-5}$. The Dice term is particularly useful because the occupancy maps are often sparse and highly imbalanced.

The region loss supervises the structural region labels using weighted cross-entropy~(WCE):
\[
L_R=\mathrm{WCE}(R'_{\mathrm{pred}},R).
\]
We use five classes, namely \emph{gap}, \emph{wall}, \emph{infill}, \emph{support}, and \emph{skirt}. In implementation, the class weights are set to
\[
[0.1,\ 3.0,\ 1.0,\ 2.0,\ 0.5],
\]
which down-weights the dominant \emph{gap} class and emphasizes relatively sparse but important classes such as \emph{wall} and \emph{support}.

The flow loss supervises the predicted extrusion flow only in printed regions. Following the main manuscript, we use a masked Huber~\cite{collins1976robust} loss:
\[
L_Q=\mathrm{Huber}\!\left(Q'_{\mathrm{pred}},Q;M\right),
\]
where the occupancy mask $M$ restricts supervision to printed pixels. More explicitly, if $\delta$ denotes the Huber threshold and $\Omega=\{i\mid M_i=1\}$ denotes the set of occupied pixels, then
\[
L_Q=\frac{1}{|\Omega|}\sum_{i\in\Omega}\mathrm{Huber}_\delta\!\left(Q'_{\mathrm{pred},i}-Q_i\right).
\]
In addition, the occupancy and region predictions are structurally coupled in the decoder: occupancy is derived from the region prediction by treating all non-gap classes as occupied. Therefore, the occupancy loss and region loss are complementary, with $L_M$ supervising the coarse printed-vs-empty layout and $L_R$ providing finer semantic guidance within the printed area.

Overall, the loss formulation jointly supervises geometry, semantic region structure, and extrusion flow, while the slice sampling strategy keeps training efficient for objects with many layers.

% \section{Additional comparisons with state-of-the-art methods}
% \label{sec:supp_sota_comp}

\section{Detailed computation of strength and cost proxies}
\label{sec:supp_metrics}

This section provides the detailed computation of the strength and cost proxies. 
For each layer $\ell$, let $M_\ell \in \mathbb{R}^{H\times W}$ denote the occupancy map and $R_\ell \in \mathbb{R}^{H\times W}$ denote the region map. 
A pixel is considered occupied when $M_\ell(i,j)>0$, and an infill pixel is defined as a pixel assigned to the infill region in $R_\ell$. 
All quantities are first computed per layer and then aggregated across layers.
The strength proxy combines three factors: infill material ratio, infill connectivity, and directional balance. 
For each layer, we compute the number of occupied pixels
\[
\mathrm{occ}_\ell=\#\{(i,j)\mid M_\ell(i,j)>0\},
\]
and the number of infill pixels
\[
\mathrm{inf}_\ell=\#\{(i,j)\mid R_\ell(i,j)\in \mathcal{I}\},
\]
where \(\mathcal{I}\) denotes the infill region. Summing over all layers gives
\[
\mathrm{occ}_{\mathrm{tot}}=\sum_\ell \mathrm{occ}_\ell,\qquad
\mathrm{inf}_{\mathrm{tot}}=\sum_\ell \mathrm{inf}_\ell.
\]
The infill material ratio is then
\[
r_{\mathrm{inf}}=\frac{\mathrm{inf}_{\mathrm{tot}}}{\max(\mathrm{occ}_{\mathrm{tot}},1)}.
\]
We next define the binary infill mask
\[
I_\ell(i,j)=\mathbf{1}[R_\ell(i,j)\in \mathcal{I}],
\]
and compute connected components on \(I_\ell\) using 4-neighborhood connectivity. 
Let \(C_\ell^{\max}\) be the size of the largest connected infill component. The connectivity score is
\[
s_{\mathrm{conn}}
=
\frac{1}{L}\sum_{\ell=1}^{L}
\frac{C_\ell^{\max}}{\max(\mathrm{inf}_\ell,1)}.
\]

To quantify directional balance, we count infill boundary transitions in the horizontal and vertical directions:
\[
t_h^\ell=\sum_{i,j}\big|I_\ell(i,j+1)-I_\ell(i,j)\big|,\qquad
t_v^\ell=\sum_{i,j}\big|I_\ell(i+1,j)-I_\ell(i,j)\big|.
\]
The per-layer direction-balance term is
\[
d_\ell=
\frac{\min(t_h^\ell,t_v^\ell)}{\max(\max(t_h^\ell,t_v^\ell),1)},
\]
and the sample-level direction balance is
\[
s_{\mathrm{dir}}=\frac{1}{L}\sum_{\ell=1}^{L} d_\ell.
\]
The final strength proxy is
\[
\mathrm{Strength}
=
0.5\,\mathrm{clip}(r_{\mathrm{inf}})
+
0.3\,\mathrm{clip}(s_{\mathrm{conn}})
+
0.2\,\mathrm{clip}(s_{\mathrm{dir}}),
\]
where \(\mathrm{clip}(x)=\max(0,\min(1,x))\).
For the cost proxy, we use the total occupied area together with infill-boundary complexity. 
The total number of infill-boundary transitions is
\[
T_{\mathrm{tot}}=\sum_{\ell=1}^{L}(t_h^\ell+t_v^\ell),
\]
and the cost proxy is defined as
\[
\mathrm{Cost}=\mathrm{occ}_{\mathrm{tot}}+0.5\,T_{\mathrm{tot}}.
\]
Finally, the combined score is
\[
\mathrm{Combined}=10^{5}\cdot \frac{\mathrm{Strength}}{\mathrm{Cost}}.
\]
This score favors patterns that achieve high structural strength while maintaining low fabrication cost.

\input{fig/ShapeNet}

\section{Qualitative results on unseen data}
\label{sec:supp_more_qual_unseen}

We present additional qualitative results of PrintAnything on unseen point cloud data from the ShapeNet~\cite{chang2015shapenet} dataset.
In Figure~\ref{fig:supp_qual_shapenet}, we visualize four examples including airplane, bench, gun, and chair shapes.
Even for complex geometries such as airplanes and guns, the final 3D printed outputs remain intricate and preserve the geometry of the input point clouds.
These results indicate that PrintAnything generalizes well to unseen point cloud shapes that were not observed during training.

\section{Visualization of predicted G-plan map}
\label{sec:supp_vis_gplan_map}
Figure~\ref{fig:G-map} visualizes the predicted geometric plan (G-plan) maps produced by PrintAnything together with the corresponding input point clouds. 
The G-plan representation consists of an occupancy map~$M'_{\mathrm{pred}}$, a region map~$R'_{\mathrm{pred}}$, and a flow map~$Q'_{\mathrm{pred}}$. 
The region map encodes printing semantics including \emph{gap}, \emph{wall}, \emph{infill}, \emph{support}, and \emph{skirt}, which are visualized in black, red, green, blue, and yellow, respectively. 
As shown in the figure, PrintAnything predicts precise occupancy masks and region assignments that capture the geometric structure of each slice.
The predicted flow map further provides per-pixel material deposition values in the range $[0,255]$, which are used to control extrusion during layer-by-layer fabrication. 
Even for shapes with intricate geometry, such as the cogwheel example in the second row, the model accurately recovers detailed occupancy and region structures. 
In the fourth row, the predicted G-plan map correctly represents the infill pattern inside the automotive component. 
Furthermore, for thin structures such as the propeller slices, PrintAnything still produces accurate occupancy, region, and flow predictions, demonstrating the robustness of the proposed representation for complex and fine-grained geometries.

\section{Visualization of infilled G-plan map}
\label{sec:supp_vis_recomm_gplan_map}

Figure~\ref{fig:G-map_recommend} visualizes the infilled geometric plan (G-plan) maps generated by our infill recommender. 
Unlike the region map $R'_{\mathrm{pred}}$ shown in Figure~\ref{fig:G-map}, which represents the infill area as a single semantic region, the infilled region map $R_{\mathrm{pred}}$ explicitly incorporates the selected infill pattern within the infill region. 
This representation converts the predicted semantic infill region into a structured printing pattern that can be directly executed during fabrication.
Each row in the figure illustrates different infill patterns recommended by our system, including grid, cubic, gyroid, and honeycomb structures. 
From top to bottom, the rows show grid, cubic, gyroid, honeycomb, grid, and gyroid infill patterns.
The recommender selects an appropriate infill strategy for each shape to balance the trade-off between structural strength and material cost. 
As shown in the examples, the recommended infill patterns are diverse yet structurally consistent with the geometry of the printed object.

% \section{Additional results on incomplete and noisy point cloud}
% \label{sec:supp_more_incomplete_noisy}

% \section{Additional qualitative results}
% \label{sec:supp_more_qual}

\input{fig/erroneous_mesh}

\section{Results of erroneous mesh reconstruction}
\label{sec:supp_error_mesh}

Figure~\ref{fig:err_mesh} visualizes erroneous meshes produced by previous methods~\cite{kazhdan2006poisson, chenmeshanything, liu2025diffusing} on Slice-100K~\cite{jignasu2024slice} dataset that reconstruct 3D meshes from input point clouds. 
We observe that these methods often generate meshes with artifacts such as holes and inverted faces, even for simple shapes with dense and clean input point clouds. 
In particular, point-cloud-to-mesh reconstruction can be challenging even for simple planar surfaces. 
These examples demonstrate that mesh reconstruction frequently fails even in seemingly straightforward cases, which supports the motivation behind PrintAnything, as erroneous mesh reconstruction often leads to unsuccessful 3D printing results. 
In contrast, PrintAnything directly operates on input point clouds and predicts 3D printing G-code that can be converted into printable shapes, thereby avoiding the failure modes of mesh reconstruction and producing robust and accurate 3D printing outputs.

\input{tab/supp_inferencetime}

\section{Running time comparison}
\label{sec:supp_efficiency}

Table~\ref{tab:inference_time_comparison} presents the average inference time for 3D printing G-code generation across different methods.
We compare our approach with traditional geometry-based reconstruction using Poisson surface reconstruction~\cite{kazhdan2006poisson}, learning-based mesh generation with MeshAnything~\cite{chenmeshanything}, and a recent method DWG~\cite{liu2025diffusing}.
The results show that our method achieves the fastest inference time, requiring only $0.33$ seconds on average, which is slightly faster than DWG ($0.37$ seconds) and much faster than Poisson reconstruction and MeshAnything.
This significant efficiency gain comes from directly predicting printable G-code without requiring expensive mesh reconstruction or optimization procedures.
The results demonstrate that our approach enables efficient and practical G-code generation suitable for real-time or interactive 3D printing applications.

\section{Limitations and societal impacts}
\label{sec:supp_limit_and_societal}

\noindent\textbf{Limitations.}
PrintAnything addresses two major challenges in 3D printing from unoriented point clouds: existing slicers do not directly support point-cloud inputs, and erroneous mesh reconstruction often leads to printing failures.
By directly predicting printable structures from point clouds, our framework enables robust G-code generation without relying on intermediate mesh reconstruction. 
Despite these advances, our current formulation does not explicitly incorporate physics inductive bias~\cite{de2018end, greydanus2019hamiltonian, wright2022deep, mezghanni2022physical} of the 3D printing process. 
Although the proposed geometric plan map produces stable and printable G-code in practice, the model primarily relies on geometric reasoning and does not explicitly account for material behavior or physical constraints during fabrication. 
Furthermore, the availability of large-scale 3D printing datasets remains inherently limited, making supervised learning difficult to scale.
Exploring self-supervised learning strategies~\cite{hong2023acl, hong2025starrygazer} for reducing the reliance on annotated training data is necessary.

\noindent\textbf{Societal impacts.}
The proposed method has the potential to benefit a wide range of applications in additive manufacturing and digital fabrication, particularly in scenarios where printable structures must be generated directly from point cloud data.
Such capabilities may support rapid prototyping, manufacturing automation, and broader accessibility of 3D printing technologies for users without complete CAD models.
However, as with many fabrication technologies, there is also a potential risk of misuse.
In particular, automated generation of printable G-code from geometric inputs could be exploited to produce unauthorized or harmful objects, such as weapons or other dangerous items that may pose risks to public safety.
Therefore, the development and deployment of PrintAnything should be accompanied by careful consideration of ethical and societal implications.
Responsible use in accordance with applicable regulations and safety guidelines is encouraged, and future work may explore safety-aware constraints or filtering strategies to further mitigate potential misuse.

\include{fig/G-plan_map}

\include{fig/G-plan_map_recommend}

%% file: algo/slice-wise-conversion.tex
\begin{algorithm}[h!]
\caption{Slice-wise conversion from G-plan map to G-code}
\label{alg:gplan_to_gcode}
\begin{algorithmic}[1]
\Require Layer-wise G-plan map with occupancy mask $M$, region map $R$, and flow map $Q$
\Require Printer settings: line width $w$, layer height $h$, number of walls $K$, infill density $\rho$, and print speeds
\Ensure G-code with perimeters and infill

\State Load G-plan map and initialize G-code file
\State Compute filament area $A_f = \pi (d_{\mathrm{fil}}/2)^2$
\State Write printer header

\For{each layer}
    \State Load $M$, $R$, and $Q$
    \State Write layer header comment \\

    % \Comment{Generate perimeters}
    \State \textbf{Perimeter generation}
    \State Initialize working mask $W \gets M$
    \For{$k = 1$ to $K$}
        \State Extract contours from $W$
        \If{no contour exists}
            \State \textbf{break}
        \EndIf
        \For{each contour}
            \State Simplify contour and convert vertices to printer coordinates
            \State Assign region-specific print attributes using $R$
            \State Move to the first point
            \For{each contour segment of length $\ell$}
                \State $E_{\mathrm{base}} \gets \dfrac{w h \ell}{A_f}$
                \State Sample $q_v$ from $Q$ at the segment midpoint
                \State $E \gets E_{\mathrm{base}} \cdot q_v$
                \State Write extrusion move
            \EndFor
        \EndFor
        \State Erode $W$ to obtain the next inner wall
    \EndFor\\

    % \Comment{Generate infill}
    \State \textbf{Infill generation}
    \State Compute scanline spacing from infill density $\rho$
    \For{each selected scanline}
        \State Find occupied intervals in $M$
        \For{each interval of length $\ell$}
            \State Assign region-specific print attributes using $R$
            \State Move to interval start
            \State $E_{\mathrm{base}} \gets \dfrac{w h \ell}{A_f}$
            \State Sample $q_v$ from $Q$ at the interval midpoint
            \State $E \gets E_{\mathrm{base}} \cdot q_v$
            \State Write extrusion move to interval end
        \EndFor
    \EndFor
\EndFor

\State Write G-code footer
\end{algorithmic}
\end{algorithm}

%% file: algo/gcode-to-gplan1.tex
\begin{algorithm}[h!]
\caption{G-code to G-plan map rasterization (Part 1)}
\label{alg:gcode_to_gplan}
\begin{algorithmic}[1]
\Require Single-extruder FDM G-code file
\Require Optional rasterization settings: pixel size $p$, filament diameter $d_f$, nozzle width $d_n$
\Ensure Per-layer G-plan maps $(M,R,Q)$ and a manifest file

\State Initialize parser state and empty layer list

\State \textbf{Parse deposited toolpath segments}
\For{each line in the G-code file}
    \State Update parser state and layer metadata
    \If{the line corresponds to an extruding motion}
        \State Convert it into a deposited segment with endpoints, layer height, width, speed, and type
        \State Append the segment to the current layer
    \EndIf
\EndFor

\algstore{gcodetogplan}
\end{algorithmic}
\end{algorithm}

%% file: algo/gcode-to-gplan2.tex
\begin{algorithm}[h!]
\ContinuedFloat
\caption{G-code to G-plan map rasterization (Part 2)}
\begin{algorithmic}[1]
\algrestore{gcodetogplan}

\State \textbf{Estimate printing parameters}
\State Determine $d_f$ and $d_n$ from the G-code header if available
\State If $p$ is not provided, choose it from $d_n$

\State \textbf{Compute shared raster bounds}
\State Collect all segment endpoints across layers
\State Compute a global 2D bounding box and raster grid shared by all layers

\State \textbf{Rasterize each layer into the G-plan map}
\State $A_f \gets \pi (d_f/2)^2$

\For{each layer}
    \State Initialize occupancy mask $M$, region map $R$, and flow map $Q$ as zero arrays
    \For{each deposited segment in the layer}
        \State Determine segment width, height, region label, and volumetric flow
        \State Rasterize the segment onto the grid using its local width
        \For{each covered pixel}
            \State Set $M$ to occupied
            \State Assign or update the region label in $R$
            \State Accumulate the segment flow into $Q$
        \EndFor
    \EndFor
    \State Normalize $Q$ over covered pixels
    \State Save $M$, $R$, and $Q$ for the current layer
\EndFor

\State \textbf{Write manifest}
\State Save raster metadata, printer/material parameters, and per-layer file paths
\end{algorithmic}
\end{algorithm}

%% file: fig/ShapeNet.tex
\begin{figure}[t]
\centering
    \includegraphics[width=\linewidth]{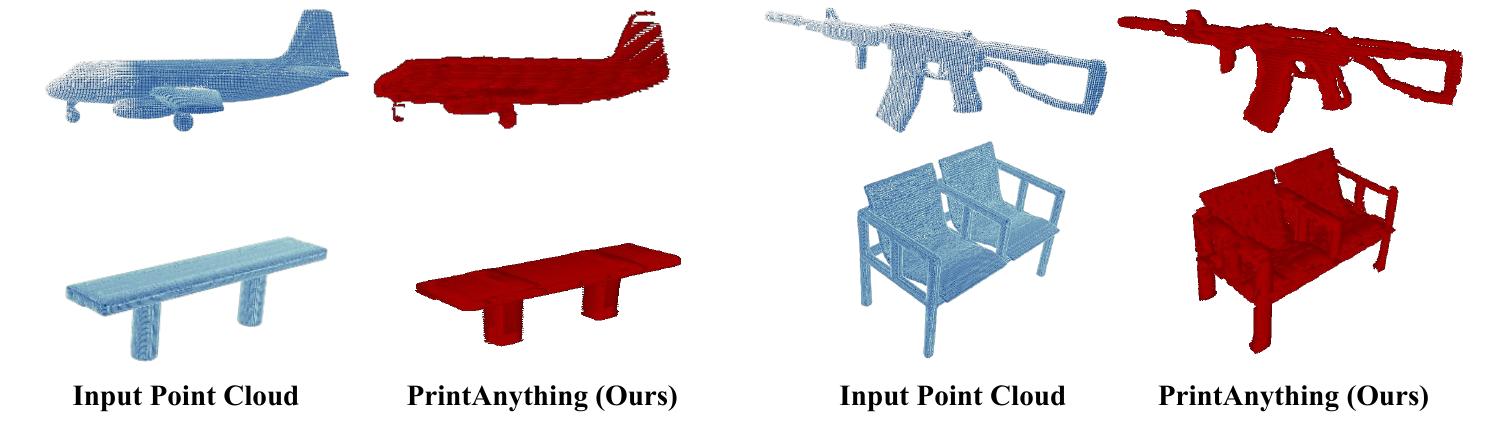}
    \caption{
        \textbf{Qualitative Results on ShapeNet~\cite{chang2015shapenet} dataset.} 
    }
    \label{fig:G-map}
\label{fig:supp_qual_shapenet}
\end{figure}

%% file: fig/erroneous_mesh.tex
\begin{figure*}[t]
\centering
    \includegraphics[width=\linewidth]{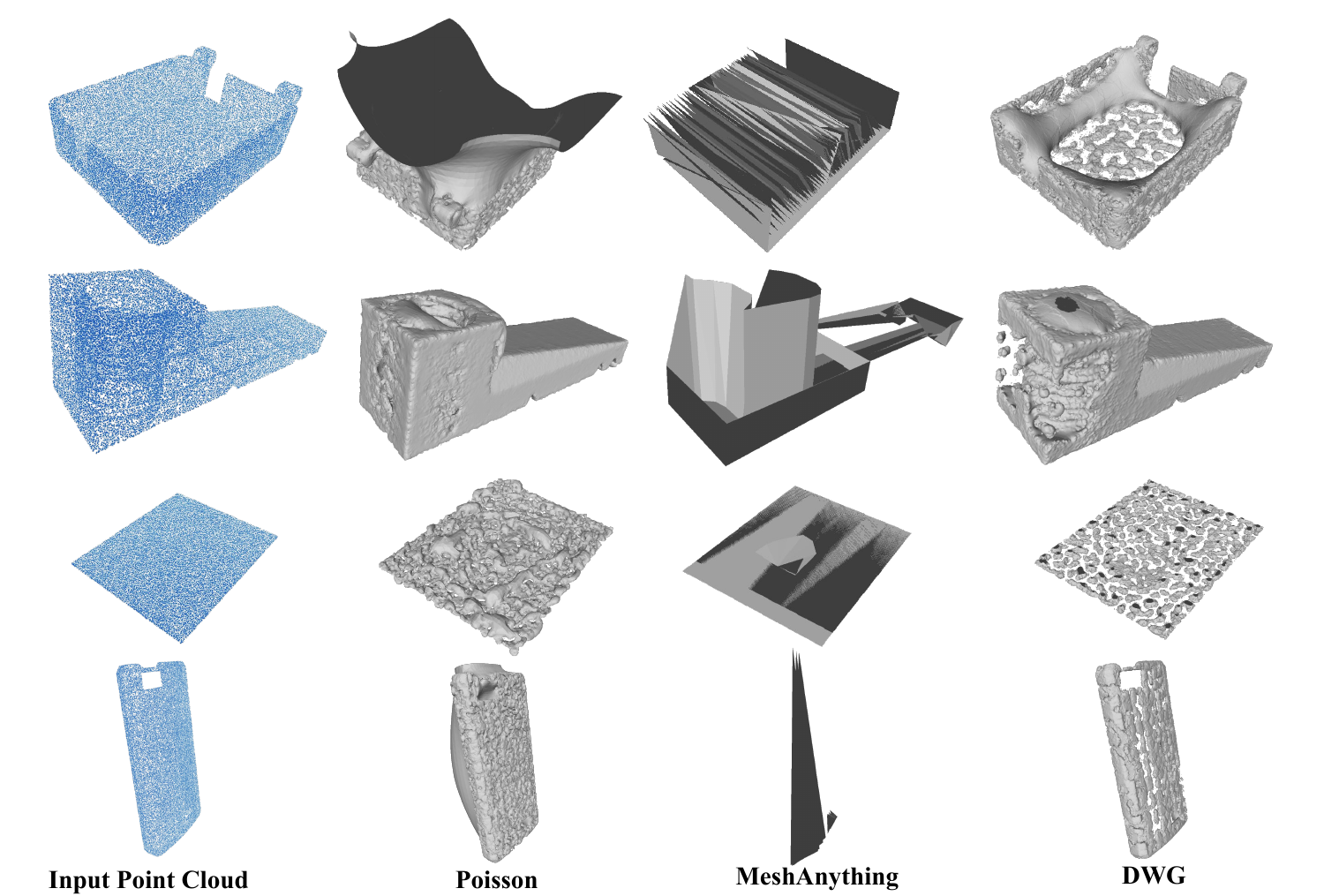}
    \caption{
        \textbf{Visualization of erroneous mesh on Slice-100K~\cite{jignasu2024slice} dataset.} 
    }
    \label{fig:err_mesh}
\end{figure*}

%% file: tab/supp_inferencetime.tex
\begin{table}[h]
\centering
\caption{Average inference time comparison of 3D printing G-code generation.}
\label{tab:inference_time_comparison}
\begin{tabular}{l c}
\toprule
\textbf{Method} & \textbf{Average Inference Time~(sec)} \\
\midrule
Poisson surface reconstruction~\cite{kazhdan2006poisson} & 123.97 \\
MeshAnything~\cite{chenmeshanything} & 113.62 \\
DWG~\cite{liu2025diffusing} & 0.37 \\
Ours & \textbf{0.33} \\
\bottomrule
\end{tabular}
\end{table}

%% file: fig/G-plan_map.tex
\begin{figure*}[t]
\centering
    \includegraphics[width=\linewidth]{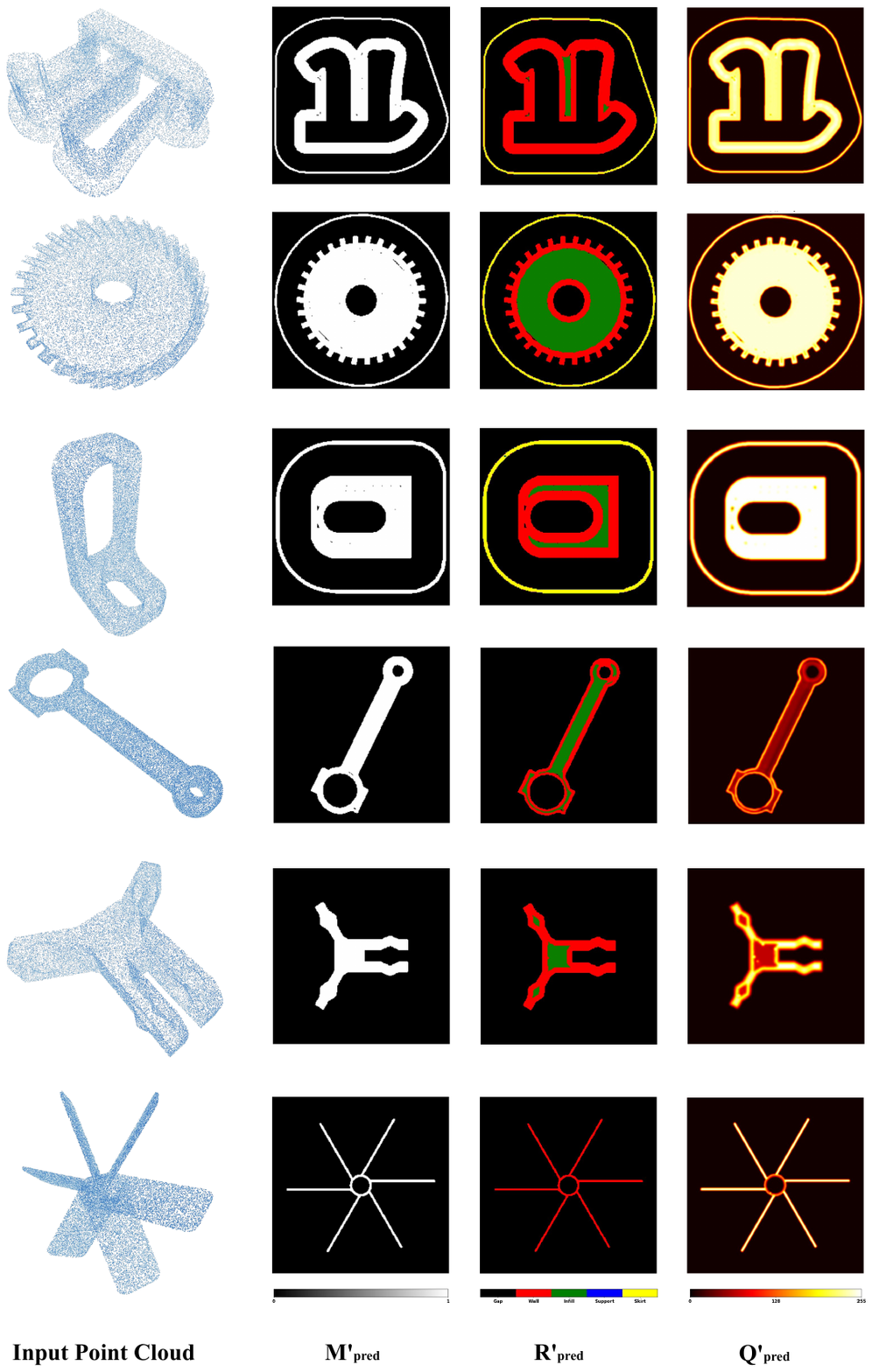}
    \caption{
        \textbf{Visualization of predicted G-plan map on Slice-100K~\cite{jignasu2024slice} dataset.} 
    }
    \label{fig:G-map}
\end{figure*}

%% file: fig/G-plan_map_recommend.tex
\begin{figure*}[t]
\centering
    \includegraphics[width=\linewidth]{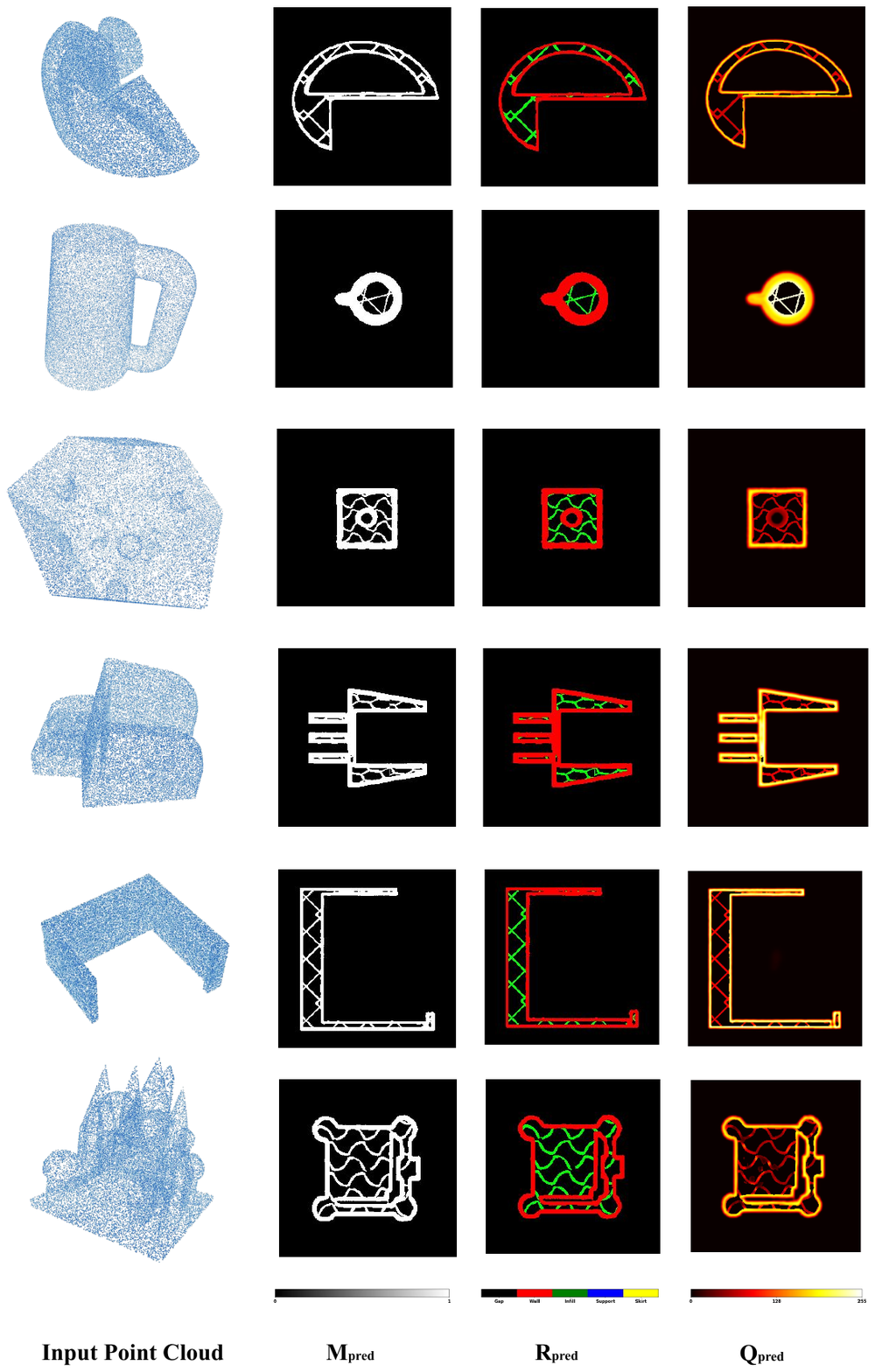}
    \caption{
        \textbf{Visualization of infilled G-plan map on Slice-100K~\cite{jignasu2024slice} dataset.} 
    }
    \label{fig:G-map_recommend}
\end{figure*}